\documentclass[11pt]{article}

\usepackage[in]{fullpage}
\usepackage{times}
\usepackage{graphicx}
\usepackage[rflt]{floatflt}
\usepackage{epsfig,subfigure,color,multirow}
\usepackage{amsmath,amssymb,algorithm,algorithmic,theorem,float,bbm,bm,enumerate}
\usepackage{url}
\usepackage{epstopdf}

\usepackage{times}
\usepackage{graphicx}
\usepackage[rflt]{floatflt}
\usepackage{epsfig,subfigure}
\usepackage{amsmath,amssymb,algorithm,algorithmic,theorem,float,bbm,bm,enumerate,multirow}
\usepackage{rotating}
\usepackage{array}

\usepackage[small,it]{caption}
\usepackage[small,compact]{titlesec}
\usepackage{times,verbatim}
\usepackage{color}
\usepackage{url}

\usepackage{chngcntr}

\DeclareMathOperator*{\argmin}{argmin}
\DeclareMathOperator*{\sign}{sign}

\newcommand{\beq}{\begin{equation}}
\newcommand{\eeq}{\end{equation}}

\newcommand{\bes}{\begin{split}}
\newcommand{\ees}{\end{split}}

\newcommand{\benum}{\begin{enumerate}}
\newcommand{\eenum}{\end{enumerate}}

\newcommand\R{\mathbb{R}}
\renewcommand\P{\mathbb{P}}
\newcommand\C{\mathbb{C}}

\newcommand\s{\mathbb{S}}
\newcommand\I{\mathbb{I}}

\renewcommand{\t}{\mathbf{t}}

\renewcommand{\u}{\mathbf{u}}
\renewcommand{\v}{\mathbf{v}}
\newcommand{\w}{\mathbf{w}}
\newcommand{\x}{\mathbf{x}}
\newcommand{\y}{\mathbf{y}}
\newcommand{\z}{\mathbf{z}}
\newcommand{\g}{\mathbf{g}}

\newcommand{\cC}{{\cal C}}

\newcommand{\cL}{{\cal L}}

\newcommand{\cR}{{\cal R}}
\newcommand{\cT}{{\cal T}}

\newcommand{\cG}{{\cal G}}

\newcommand{\cN}{{\cal N}}

\newcommand{\cA}{\mathcal{A}}

\newcommand{\bA}{\mathbf{A}}

\newcommand{\bX}{\mathbf{X}}

\newcommand{\htheta}{{\hat{\btheta}}}
\newcommand{\hDelta}{{\hat{\Delta}}}

\newcommand{\E}{\mathbf{E}}

\newcommand{\btheta}{\boldsymbol{\theta}}

\newcommand{\myref}[1]{(\ref{#1})}

\newcounter{exampleI}
{\theorembodyfont{\rmfamily} \theoremstyle{plain} }

\newcounter{exampleII}
{\theorembodyfont{\rmfamily} \theoremstyle{plain} }

\newcounter{exampleIII}
{\theorembodyfont{\rmfamily} \theoremstyle{plain} }

{\theorembodyfont{\rmfamily} }
{\theorembodyfont{\rmfamily} }
\newtheorem{theo}{Theorem}
\newtheorem{prop}{Proposition}
\newtheorem{lemm}{Lemma}
\newtheorem{corr}{Corollary}

\newcommand{\proof}{\noindent{\itshape Proof:}\hspace*{1em}}

\newcommand{\qed}{\nolinebreak[1]~~~\hspace*{\fill} \rule{5pt}{5pt}\vspace*{1mm}\vspace*{1mm}}

\newcommand {\commentout}[1] {}

\newcommand{\ksup}[1]{\|#1\|_k^{sp}}
\newcommand{\ksupd}[1]{\|#1\|_k^{sp^*}}
\newcommand{\ceil}[1]{\left\lceil#1\right\rceil}

\title{Generalized Dantzig Selector:\\
Application to the k-support norm}

\author{Soumyadeep Chatterjee \\
Dept of Computer Science \& Engg\\
University of Minnesota, Twin Cities\\
chatter@cs.umn.edu
\and
Sheng Chen \\
Dept of Computer Science \& Engg\\
University of Minnesota, Twin Cities\\
shengc@cs.umn.edu
\and
Arindam Banerjee\\
Dept of Computer Science \& Engg\\
University of Minnesota, Twin Cities\\
banerjee@cs.umn.edu
}

\date{}

\begin{document}

\maketitle

\counterwithout{equation}{section}


\begin{abstract}
We propose a Generalized Dantzig Selector (GDS) for linear models, in which any norm encoding the parameter structure can be leveraged for estimation. We investigate both computational and statistical aspects of the GDS. Based on conjugate proximal operator, a flexible inexact ADMM framework is designed for solving GDS, and non-asymptotic high-probability bounds are established on the estimation error, which rely on Gaussian width of unit norm ball and suitable set encompassing estimation error. Further, we consider a non-trivial example of the GDS using $k$-support norm. We derive an efficient method to compute the proximal operator for
$k$-support norm since existing methods are inapplicable in this setting. For statistical analysis, we provide upper bounds for the Gaussian widths needed in the GDS analysis, yielding the first statistical recovery guarantee for estimation with the $k$-support norm. The experimental results confirm our theoretical analysis. 
\end{abstract}

\section{Introduction}

The Dantzig Selector (DS)~\cite{birt09,ct07} provides an alternative to regularized regression approaches such as Lasso~\cite{tibs96,zhyu06} for sparse estimation. While DS does not consider a regularized maximum likelihood approach, \cite{birt09} has established clear similarities between the estimates from DS and Lasso. While norm regularized regression approaches have been generalized to more general norms, such as decomposable norms~\cite{nrwy12}, the literature on DS has primarily focused on the sparse $L_1$ norm case, with a few notable exceptions which have considered extensions to sparse group-structured norms~\cite{ljjl10}.

In this paper, we consider linear models of the form $\y = \bX\btheta^* + \w$,
where  $\y \in \R^n$ is a set of observations, $\bX\in \R^{n\times p}$ is a design matrix, and $\w \in \R^n $ is i.i.d.~noise. For {\em any} given norm $\cR(\cdot)$, the parameter $\btheta^*$ is assumed to structured in terms of having a low value of $\cR(\btheta^*)$. For this setting, we propose the following Generalized Dantzig Selector (GDS) for parameter estimation:
\beq
\begin{split}
\hat{\btheta}& = \argmin_{\btheta \in \R^p}~ \cR(\btheta)\\
\text{ s.t.}&~\cR^*\big(\bX^T(\y-\bX \btheta)\big) \leq \lambda_p ~,
\label{eq:dantzig minimization}
\end{split}
\eeq
where $\cR^*(\cdot)$ is the dual norm of $\cR(\cdot)$, and $\lambda_p$ is a suitable constant.
If $\cR(\cdot)$ is the $L_1$ norm, \eqref{eq:dantzig minimization} reduces to standard DS~\cite{ct07}. A key novel aspect of GDS is that the constraint is in terms of the dual norm $\cR^*(\cdot)$ of the original structure inducing norm $\cR(\cdot)$. It is instructive to contrast GDS with the recently proposed atomic norm based estimation framework~\cite{crpw12} which, unlike GDS, considers constraints based on the $L_2$ norm of the error $\| \y-\bX\btheta\|_2$, and focuses only on atomic norms.

In this paper, we consider both computational and statistical aspects of the GDS.
For the $L_1$-norm Dantzig selector, \cite{ct07} proposed a primal-dual interior point method since the optimization is a linear program. DASSO and its generalization proposed in \cite{jrl09, jr09} focused on homotopy methods, which provide a piecewise linear solution path through a sequential simplex-like algorithm. However, none of the algorithms above can be immediately extended to our general formulation.
In recent work, the Alternating Direction Method of Multipliers (ADMM) has been applied to the $L_1$ Dantzig selection problem~\cite{lpz12,wy12}, and the linearized version in~\cite{wy12} proved to be efficient. Motivated by such results for DS, we propose a general inexact ADMM~\cite{wb13} framework for GDS where the primal update steps, interestingly, turn out respectively to be proximal updates involving $\cR(\btheta)$ and its convex conjugate, the indicator of $\cR^*(\x) \leq \lambda$. As a result, by Moreau decomposition, it suffices to develop efficient proximal update for either $\cR(\btheta)$ or its conjugate.
On the statistical side, we establish non-asymptotic high-probability bounds on the estimation error $\|\hat{\btheta} - \btheta^* \|_2$. Interestingly, the bound depends on the Gaussian width of the unit norm ball of $\cR(\cdot)$ as well as the Gaussian width of suitable set where the estimation error belongs~\cite{crpw12,rarn12}.




As a non-trivial example of the GDS framework, we consider estimation using the recently proposed $k$-support norm~\cite{afs12,mps14}. We show that proximal operators for $k$-support norm can be efficiently computed in $O(p \log p + \log k \log (p-k))$, and hence the estimation can be done efficiently. Note that existing work~\cite{afs12,mps14} on $k$-support norm has focused on the proximal operator for the {\em square} of the $k$-support norm, which is not directly applicable in our setting. On the statistical side, we provide upper bounds for the Gaussian widths of the unit norm ball and the error set as needed in the GDS framework, yielding the first statistical recovery guarantee for estimation with the $k$-support norm.


The rest of the paper is organized as follows: We establish general optimization and statistical recovery results for GDS for any norm in Section~\ref{sec:general}. In Section~\ref{sec:ksupp}, we present efficient algorithms and estimation error bounds for the $k$-support norm. We present experimental results in Section~\ref{sec:expt} and conclude in Section~\ref{sec:conc}. All technical analyses and proofs are in the supplement.

\section{General Optimization and Statistical Recovery Guarantees}
\label{sec:general}
The problem in~\eqref{eq:dantzig minimization} is a convex program, and a suitable choice of $\lambda_p$ ensures that the feasible set is not empty. We start the section with an inexact ADMM framework for solving problems of the form~\eqref{eq:dantzig minimization}, and then present bounds on the estimation error establishing statistical consistency of GDS.

\subsection{General Optimization Framework using Inexact ADMM}
In optimization, we temporarily drop the subscript $p$ of $\lambda_p$ for convenience. We let $\bA = \bX^T \bX$, $\u = \bX^T \y$, and define the set $\cC_{\lambda} = \{ \v ~:~ \cR^*(\v) \leq \lambda \}$. The optimization problem is equivalent to
\begin{equation}
\label{equiform}
\min_{\btheta, \v \in \R^p} \ \cR(\btheta) ~\text{~~~~s.t.} \ \ \u - \bA \btheta = \v, \ \v \in \cC_{\lambda} ~.
\end{equation}
Due to the nonsmoothness of both $\cR$ and $\cR^*$, solving \eqref{equiform} can be quite challenging and a generally applicable algorithm is Alternating Direction Method of Multipliers (ADMM). The augmented Lagrangian function for \eqref{equiform} is given as
\begin{equation}
\label{augLagrangian}
\mathcal{L}_{\cR}(\btheta, \v, \z) = \cR(\btheta) + \langle \z,  \bA \btheta + \v - \u \rangle + \frac{\rho}{2} || \bA \btheta  + \v - \u||_2^2 ~.
\end{equation}
in which $\z$ is the Lagrange multiplier and $\rho$ controls the penalty introduced by the quadratic term. The iterative updates of the variables $(\btheta, \v, \z)$ in standard ADMM are given by
\begin{align}
\label{thetaupdate}
&\btheta^{k+1} \gets \argmin_{\btheta} \mathcal{L}_{\cR}(\btheta, \v^k, \z^k) ~, \\
\label{xupdate}
&\v^{k+1} \gets \argmin_{\v \in \cC_{\lambda}} \mathcal{L}_{\cR}(\btheta^{k+1}, \v, \z^k) ~,\\
\label{zupdate}
&\z^{k+1} \gets \z^k + \rho ( \bA \btheta^{k+1} + \v^{k+1} - \u) ~.
\end{align}
Note that update \eqref{thetaupdate} amounts to a regularized least squares problem of $\btheta$, which can be computationally expensive. Thus we use an inexact update for $\btheta$ instead, which can alleviate the computational cost and lead to a quite simple algorithm. Inspired by~\cite{wy12}, we consider a simpler subproblem for the $\btheta$-update which minimizes
\begin{equation}
\label{LinearizedLagrangian}
\begin{split}
\widetilde{\mathcal{L}}_{\cR}^k(\btheta, \v^k, \z^k) &= \cR(\btheta) + \langle \z^k, \bA \btheta + \v^k - \u \rangle +
\frac{\rho}{2} \Big ( \big \| \bA \btheta^k + \v^k - \u \big \|_2^2 + \\
&  2 \big \langle \btheta - \btheta^k, \bA^T(\bA \btheta^{k} + \v^k - \u) \big \rangle + \frac{\mu}{2} \big \| \btheta- \btheta^k \big \|_2^2 \Big ) ~,
\end{split}
\end{equation}
where $\mu$ is a user-defined parameter. $\widetilde{\mathcal{L}}_{\cR}^k(\btheta, \v^k, \z^k)$ can be viewed as an approximation of $\mathcal{L}_{\cR}(\btheta, \v^k, \z^k)$ with the quadratic term linearized at $\btheta^k$. Then the update \eqref{thetaupdate} is replaced by
\begin{equation}
\label{update_theta}
\begin{split}
\btheta^{k+1} &\gets \argmin_{\btheta} \widetilde{\mathcal{L}}_{\cR}^k(\btheta, \v^k, \z^k) \\
&= \argmin_{\btheta}  \bigg \lbrace \frac{2 \cR(\btheta)}{\rho \mu} + \frac{1}{2} \Big \|\btheta - \big (\btheta^k - \frac{2}{\mu} \bA^T(
\bA \btheta^k + \v^k - \u + \frac{\z^k}{\rho}) \big ) \Big \|_2^2 \bigg \rbrace ~.
\end{split}
\end{equation}
Similarly the update of $\v$ in \eqref{xupdate} can be recast as
\begin{equation}
\label{update_x}
\v^{k+1} \gets \argmin_{\v \in \cC_{\lambda}} \mathcal{L}_{\cR}(\btheta^{k+1}, \v, \z^k) = \argmin_{\v \in \cC_{\lambda}} \frac{1}{2} \big \|\v - (\u - \bA \btheta^{k+1} - \frac{\z^k}{\rho}) \big \|_2^2 ~.
\end{equation}
In fact, the updates of both $\btheta$ and $\v$ turn out to compute certain \emph{proximal operators}. In general, the proximal operator $\mathbf{prox}_{h}(\cdot)$ of a closed proper convex function $h : \R^p \to \R \cup \{ + \infty \}$ is defined as
\begin{align*}
\mathbf{prox}_{h}(\x) = \argmin_{\w \in \R^p} \Big \{ \frac{1}{2} \|\w - \x \|_2^2 + h(\w) \Big \} ~.
\end{align*}
Hence it is easy to see that \eqref{update_theta} and \eqref{update_x} correspond to $\mathbf{prox}_{\frac{2\mathcal{R}}{\rho \mu}}(\cdot)$ and $\mathbf{prox}_{\mathbbm{I}_{\cC_{\lambda}}}(\cdot)$, respectively, where $\mathbbm{I}_{\cC_{\lambda}}(\cdot)$ is the indicator function of set $\cC_{\lambda}$ given by
\begin{align*}
\mathbbm{I}_{\cC_{\lambda}}(\x) = \left \{
             \begin{array}{lll}
              0 \ \ \ &\text{if \ $\x \in \cC_{\lambda}$} \\
              + \infty \ \ \ &\text{if \ otherwise}
             \end{array}  \right. ~.
\end{align*}
In Algorithm \ref{AlgADMM}, we provide our general ADMM for the GDS.
\begin{algorithm}[t!]
\renewcommand{\algorithmicrequire}{\textbf{Input:}}
\renewcommand{\algorithmicensure} {\textbf{Output:} }
\caption{ADMM for Generalized Dantzig Selector}
\label{AlgADMM}
\begin{algorithmic}[1]
\REQUIRE ~$\bA = \bX^T \bX$, $\u = \bX^T\y$, $\rho$, $\mu$ \\
\ENSURE ~Optimal $\htheta$ of \eqref{eq:dantzig minimization} \\
\STATE Initialize $(\btheta, \v, \z)$
\WHILE {not converged}
\STATE $\btheta^{k+1} \gets \mathbf{prox}_{\frac{2 \cR}{\rho \mu}} \big (\btheta^k - \frac{2}{\mu} \bA^T(\bA \btheta^k + \v^k - \u +
\frac{\z^k}{\rho}) \big )$
\label{thetastep}
\STATE $\v^{k+1} \gets \mathbf{prox}_{\mathbbm{I}_{\cC_{\lambda}}} \big (\u - \bA \btheta^{k+1}  - \frac{\z^k}{\rho} \big)$
\label{xstep}
\STATE $\z^{k+1} \gets \z^k + \rho (\bA \btheta^{k+1} + \v^{k+1} - \u)$
\ENDWHILE
\end{algorithmic}
\end{algorithm}
For the ADMM to work, we need two subroutines that can efficiently compute the proximal operators for the functions in Line \ref{thetastep} and \ref{xstep} respectively.
The simplicity of the proposed approach stems from the fact that we in fact need {\em only one} subroutine, for any one of the functions, since the functions are conjugates of each other.
\begin{prop}
\label{moreaudecomp}
Given $\beta > 0$ and a norm $\cR(\cdot)$, the two functions, $f(\x) = \beta \cR(\x)$ and $g(\x) = \mathbbm{I}_{\cC_{\beta}}(\x)$ are convex
conjugate to each other, thus giving the following identity,
\begin{equation}
\label{decomposition}
\x = \mathbf{prox}_f(\x) + \mathbf{prox}_g(\x) ~.
\end{equation}
\end{prop}
\proof The Proposition \ref{moreaudecomp} simply follows the definition of convex conjugate and dual norm, and \eqref{decomposition} is just \emph{Moreau decomposition} provided in~\cite{pb14}. \qed

The decomposition enables conversion of the two types of proximal operator to each other at negligible cost (i.e., vector subtraction). Thus we have the flexibility in Algorithm~\ref{AlgADMM} to focus on the proximal operator that is efficiently computable, and the other can be simply obtained through \eqref{decomposition}.

\noindent\textbf{Remark on convergence:} ~Note that Algorithm \ref{AlgADMM} is a special case of inexact Bregman ADMM proposed in~\cite{wb13}, which matches the case of linearizing quadratic penalty term by using $B_{\varphi_{\btheta}'}(\btheta, \btheta_k) = \frac{1}{2}\| \btheta - \btheta_k \|_2^2$ as Bregman divergence. In order to converge, the algorithm requires $\frac{\mu}{2}$ to be larger than the spectral radius of $\bA^T\bA$, and the convergence rate is $O(1/T)$ according to Theorem 2 in~\cite{wb13}.

\subsection{Statistical Recovery for Generalized Dantzig Selector}

Our goal is to provide error bounds on $\|\hat{\btheta} - \btheta^*\|_2$ between the population parameter $\btheta^*$ and the minimizer $\hat{\btheta}$ of~\eqref{eq:dantzig minimization}.
Let the error vector be defined as $\hDelta = \hat{\btheta} - \btheta^*$. For any set $\Omega \subseteq \R^p$, we would measure the size of this set using its Gaussian width~\cite{ruve08,crpw12}, which is defined as $ \omega(\Omega) = \E_\g\left[\sup_{\z\in \Omega}\langle\g, \z\rangle\right]$~,
where $\g$ is a vector of i.i.d. standard Gaussian entries. We also consider the error cone $\cT_\cR(\btheta^*)$, generated by the set of possible error vectors $\Delta$ and containing the error vector $\hDelta$, defined as
\beq
\cT_\cR(\btheta^*) := \text{\upshape cone}\left\{ \Delta \in \R^p~:~\cR(\btheta^* + \Delta) \leq \cR(\btheta^*)\right\}~.
\label{equ:tan cone}
\eeq
Note that this set contains a restricted set of directions and does not in general span the entire space of $\R^p$. Further, let $\Omega_\cR :=\{\u:\cR(\u) \leq 1 \}$. With these definitions, we obtain our main result.
\begin{theo}
Suppose the design matrix $\bX$ consists of i.i.d. Gaussian entries with zero mean variance 1, and we solve the optimization problem~\eqref{eq:dantzig minimization} with
\beq
\lambda_p \geq c\E\left[ \cR^*(\bX^T\w)\right]~.
\label{eq:lambda choice}
\eeq
Then, with probability at least $(1 - \eta_1 \exp(-\eta_2 n))$, we have
\beq
\|\hat{\btheta} - \btheta^*\|_2 \leq \frac{4c \Psi_\cR \omega(\Omega_\cR)  }{\kappa_\cL \sqrt{n}}~,
\eeq
where $\omega(\cT_\cR(\btheta^*)\cap \s^{p-1})$ is the Gaussian width of the intersection of $\cT_\cR(\btheta^*)$ and the unit spherical shell $\s^{p-1}$, $\omega(\Omega_\cR)$ is the Gaussian width of the unit norm ball,  $\kappa_\cL>0$ is the gain given by
\beq
\kappa_\cL= \frac{1}{n}\left(\ell_n - \omega(\cT_\cR(\btheta^*)\cap \s^{p-1})\right)^2~, 
\eeq
$\Psi_\cR = \sup_{\Delta\in \cT_\cR}\cR(\Delta)/\|\Delta\|_2 $ is a norm compatibility factor,
$\ell_n$ is the expected length of a length $n$ i.i.d. standard Gaussian vector with $\frac{n}{\sqrt{n+1}} < \ell_n < \sqrt{n}$, and $c>1,\eta_1,\eta_2 >0$ are constants.
\label{theo:dantzig recovery}
\end{theo}

\noindent{\bf Remark:} The choice of $\lambda_p$ is also intimately connected to the notion of Gaussian width. Note that for $\bX$ i.i.d. Gaussian entries, and $\w$ i.i.d. standard Gaussian vector, 
$\bX^T \w = \|\w\|_2 \left(\bX^T \frac{\w}{\|\w\|_2}\right)= \|\w\|_2 \z$ 
where $\z$ is an i.i.d. standard Gaussian vector. Therefore,
\begin{align}
\lambda_p \geq c\E\left[ \cR^*(\bX^T\w)\right]  &= c\E_\w[\|\w\|_2]\cdot\E_{\bX}\left[ \cR^*(\bX^T \frac{\w}{\|\w\|_2})\right] \\
&= c\E_\w[\|\w\|_2] \E_{\z}\left[ \sup_{\u:~\cR(\u)\leq 1} \langle\u, \z \rangle \right] \\
&= c\ell_n \omega\left(\Omega_\cR\right)~,
\end{align}
which is a scaled Gaussian width of the unit ball of the norm $\cR(\cdot)$.

\paragraph{Example: $L_1$-norm Dantzig Selector}
When $\cR(\cdot)$ is chosen to be $L_1$ norm, the dual norm is the $L_{\infty}$ norm, and \eqref{eq:dantzig minimization} is reduced to the standard DS, given by
\begin{equation}
\label{eq:l1minimization}
\begin{split}
\hat{\btheta}& = \argmin_{\btheta \in \R^p} \|\btheta\|_1~\text{~~~~~s.t.}~\| \bX^T(\y - \bX \btheta) \|_{\infty} \leq \lambda ~.
\end{split}
\end{equation}
We know that $\mathbf{prox}_{\beta \| \cdot \|_1}(\cdot)$ is given by the elementwise soft-thresholding operation
\begin{equation}
\big [ \mathbf{prox}_{\beta \| \cdot \|_1}(\x) \big ]_i = \sign(\x_i) \cdot \max(0, |\x_i| - \beta) ~.
\end{equation}
Based on Proposition \ref{moreaudecomp}, the ADMM updates in Algorithm \ref{AlgADMM} can be instantiated as
\begin{align*}
&\btheta^{k+1} \gets \mathbf{prox}_{\frac{2 \| \cdot \|_1}{\rho \mu}} \big (\btheta^k - \frac{2}{\mu} \bA^T(\bA \btheta^k + \v^k - \u + \frac{\z^k}{\rho}) \big ) ~, \\
&\v^{k+1} \gets (\u - \bA \btheta^{k+1}  - \frac{\z^k}{\rho}) - \mathbf{prox}_{\lambda \| \cdot \|_1} \big (\u - \bA \btheta^{k+1} - \frac{\z^k}{\rho} \big ) ~, \\
&\z^{k+1} \gets \z^k + \rho (\bA \btheta^{k+1} + \v^{k+1} - \u) ~,
\end{align*}
where the update of $\v$ leverages the decomposition \eqref{decomposition}. Similar updates were used in~\cite{wy12} for $L_1$-norm Dantzig selector.

For statistical recovery, we assume that $\btheta^*$ is $s$-sparse, i.e., contains $s$ non-zero entries, and that $\|\btheta^*\|_2 =1$, so that $\|\btheta^*\|_1 \leq s$.
It was shown in~\cite{crpw12} that the Gaussian width of the set $(\cT_{L_1}(\btheta^*)\cap \s^{p-1})$ is upper bounded as $\omega(\cT_{L_1}(\btheta^*)\cap \s^{p-1})^2 \leq 2s\log\left( \frac{p}{s}\right) + \frac{5}{4}s$. Also note that $\E\left[ \cR^*(\bX^T\w)\right] = \E[\|\w\|_2] \E[\|\g\|_\infty]\leq \sqrt{n}\sqrt{\log p}$, where $\g$ is a vector of i.i.d. standard Gaussian entries~\cite{ct07}. Further, \cite{nrwy12} has shown that $\Psi_\cR = \sqrt{s}$. Therefore, if we solve~\eqref{eq:l1minimization} with $\lambda_p = c\sqrt{n}\log p$, then
\begin{align}
\|\hat{\btheta}- \btheta^*\|_2 \leq 4c \frac{ \sqrt{s\log p} }{\kappa_\cL\sqrt{n}} = \text{\large O}\left(\sqrt{ \frac{s\log p}{n} }\right)
\end{align}
with high probability, which agrees with known results for DS~\cite{birt09,ct07}.

%


\section{Dantzig Selection with $k$-support norm}
\label{sec:ksupp}
We first introduce some notations. Given any $\btheta \in \R^p$, let $|\btheta|$ denote its absolute-valued counterpart and $\btheta^{\downarrow}$ denote the permutation of $\btheta$ with its elements arranged in decreasing order. In previous work~\cite{afs12,mps14}, the $k$-support norm is defined as
\beq
\ksup{\btheta} = \min\left\{\sum_{I\in\cG^{(k)}}\|v_I\|_2 ~:~ \text{supp}(v_I)\subseteq I,~\sum_{I\in\cG^{(k)}}v_I = \btheta \right\}~,
\eeq
where $\cG^{(k)}$ denotes the set of subsets of $\{1,\ldots,p\}$ of cardinality at most $k$. The unit ball of this norm is the set
$C_k = \text{conv}\left\{ \btheta \in\R^p~:~\|\btheta\|_0 \leq k, \|\btheta\|_2 \leq 1\right\}$~.
The dual norm of the $k$-support norm is given by
\beq
\ksupd{\btheta} = \max\left\{\|\btheta_G\|_2~:~{G\in \cG^{(k)}} \right\} =\left( \sum_{i=1}^k |\btheta|^{{\downarrow}^2}_i \right)^{\frac{1}{2}} ~.
\eeq

The $k$-support norm was proposed in order to overcome some of the empirical shortcomings of the elastic net~\cite{zoha05} and the (group)-sparse regularizers. It was shown in~\cite{afs12} to behave similarly as the elastic net in the sense that the unit norm ball of the $k$-support norm is within a constant factor of $\sqrt{2}$ of the unit elastic net ball. Although multiple papers have reported good empirical performance of the $k$-support norm on selecting highly correlated features, wherein $L_1$ regularization fails, there exists no statistical analysis of the $k$-support norm. Besides, current computational methods consider square of $k$-support norm in their formulation, which might fail to work out in certain cases.

In the rest of this section, we focus on GDS with $\cR(\btheta) = \ksup{\btheta}$ given as
\beq
\hat{\btheta} = \argmin_{\btheta \in \R^p} \|\btheta\|_k^{sp}~\text{ ~~~~~ s.t.}~\ksupd{\bX^T(\y-\bX\btheta)}\leq \lambda_p~.
\label{eq:ksup minimization}
\eeq
For the indicator function $\mathbbm{I}_{\cC_{\lambda}}(\cdot)$ of the dual norm, we present a fast algorithm for computing its proximal operator by exploiting the structure of its solution, which can be directly plugged in Algorithm~\ref{AlgADMM} to solve~\eqref{eq:ksup minimization}. Further, we prove statistical recovery bounds for $k$-support norm Dantzig selection, which hold even for a high-dimensional scenario, where $n<p$.

\subsection{Computation of Proximal Operator}

In order to solve \eqref{eq:ksup minimization}, either $\mathbf{prox}_{\lambda\| \cdot \|_k^{sp}}(\cdot)$ or $\mathbf{prox}_{\mathbbm{I}_{\cC_{\lambda}}}(\cdot)$ for $\| \cdot \|_{k}^{sp^*}$ should be efficiently computable. Existing methods~\cite{afs12,mps14} are inapplicable to our scenario since they compute the proximal operator for squared $k$-support norm, from which $\mathbf{prox}_{\mathbbm{I}_{\cC_{\lambda}}}(\cdot)$ cannot be directly obtained. In Theorem \ref{maintheorem}, we show that $\mathbf{prox}_{\mathbbm{I}_{\cC_{\lambda}}}(\cdot)$ can be efficiently computed, and thus Algorithm \ref{AlgADMM} is applicable.

\begin{theo}
\label{maintheorem}
Given $\lambda > 0$ and $\x \in \R^p$, if $\|\x\|_{k}^{sp^*} \leq \lambda$, then $\w^*=\mathbf{prox}_{\mathbbm{I}_{\cC_{\lambda}}}(\x)=\x$. If
$\|\x\|_{k}^{sp^*} > \lambda$, define $A_{sr} = \sum_{i=s+1}^r |\x|_i^{\downarrow}$, $B_{s} = \sum_{i=1}^{s}(|\x|_i^{\downarrow})^2$, in which
$0 \leq s < k$ and $k \leq r \leq p$, and construct the nonlinear equation of $\beta$,
\begin{equation}
\label{eq:nonlinearequ}
(k-s)A_{sr}^2 \left[\frac{1+\beta}{r-s+(k-s)\beta} \right]^2 - {\lambda}^2(1+\beta)^2 + B_s = 0 ~.
\end{equation}
Let $\beta_{sr}$ be given by
\begin{equation}
\beta_{sr} = \left \{
             \begin{array}{lll}
              \text{nonnegative root of \eqref{eq:nonlinearequ}} \ &\text{if $s > 0$ and the root exists} \\
              0 \ &\text{otherwise}
             \end{array}  \right. ~.
\end{equation}
Then the proximal operator $\w^*=\mathbf{prox}_{\I_{ \cC_{\lambda}}}(\x)$ is given by
\begin{equation}
\label{eq:projection}
|\w^*|_i^{\downarrow} = \left \{
             \begin{array}{lll}
             \frac{1}{1+\beta_{s^*r^*}} |\x|_i^{\downarrow} \ &\text{if} \ \ 1 \leq i \leq s^*   \\
             \sqrt{\frac{{\lambda}^2 - B_{s^*}}{k-s^*}} \ &\text{if} \ \ s^* < i \leq r^* \ \text{and} \ \beta_{s^*r^*} = 0 \\
             \frac{A_{s^*r^*}}{r^*-s^*+(k-s^*)\beta_{s^*r^*}} \ &\text{if} \ \ s^* < i \leq r^* \ \text{and} \ \beta_{s^*r^*} > 0 \\
             |\x|_i^{\downarrow} \ &\text{if} \ \ r^* < i \leq p
             \end{array}  \right. ~,
\end{equation}
where the indices $s^*$ and $r^*$ with computed $|\w^*|^{\downarrow}$ make the following two inequalities hold,
\begin{equation}
\label{eq:cond1}
|\w^*|_{s^*}^{\downarrow} > |\w^*|_{k}^{\downarrow} ~,
\end{equation}
\begin{equation}
\label{eq:cond2}
|\x|_{r^* + 1}^{\downarrow} \leq |\w^*|_{k}^{\downarrow} < |\x|_{r^*}^{\downarrow} ~.
\end{equation}
There might be multiple pairs of $(s, r)$ satisfying the inequalities \myref{eq:cond1}-\myref{eq:cond2}, and we choose the pair with the smallest $\| |\x|^{\downarrow} - |\w|^{\downarrow} \|_2$. Finally, $\w^*$ is obtained by sign-changing and reordering $|\w^*|^{\downarrow}$ to conform to $\x$.
\end{theo}

{\bf Remark:} The nonlinear equation \eqref{eq:nonlinearequ} is quartic, for which we can use general formula to get all the roots~\cite{s03}. In addition, if it exists, the nonnegative root is unique, as we show in the proof.

Theorem \ref{maintheorem} indicates that computing $\mathbf{prox}_{\mathbbm{I}_{\cC_{\lambda}}}(\cdot)$ requires sorting of entries in $|\x|$ and a two-dimensional linear search of $s^*$ and $r^*$. Hence the total time complexity is $O(p\log p + k(p-k))$. However, a more careful observation can particularly reduce the search complexity from $O(k(p-k))$ to $O(\log k\log(p-k))$, which is motivated by Theorem \ref{bisection}.

\begin{theo}
\label{bisection}
In search of ($s^*$, $r^*$) defined in Theorem \ref{maintheorem}, there can be only one $\tilde{r}$ for a given candidate $\tilde{s}$ of $s^*$, such that the inequality \eqref{eq:cond2} is satisfied. Moreover if such $\tilde{r}$ exists, then for any $r < \tilde{r}$, the associated $|\tilde{\w}|^{\downarrow}_k$ violates the first part of \eqref{eq:cond2}, and for $r > \tilde{r}$, $|\tilde{\w}|^{\downarrow}_k$ violates the second part of \eqref{eq:cond2}. On the other hand, based on the $\tilde{r}$, we have following assertion of $s^*$,
\begin{equation}
\label{s_start}
s^* \left \{
             \begin{array}{lll}
             > \tilde{s} \ &\text{if $\tilde{r}$ does not exist}  \\
             \geq \tilde{s} \ &\text{if $\tilde{r}$ exists and corresponding $|\tilde{\w}|^{\downarrow}_k$ satisfies \eqref{eq:cond1}} \\
             < \tilde{s} \ &\text{if $\tilde{r}$ exists but corresponding $|\tilde{\w}|^{\downarrow}_k$ violates \eqref{eq:cond1}}
             \end{array} \right. ~.
\end{equation}
\end{theo}

Based on Theorem \ref{bisection}, the accelerated search procedure of ($s^*$, $r^*$) is to execute a two-dimensional \emph{binary search}, and Algorithm \ref{AlgProx} gives the details. Therefore the total time complexity becomes $O(p\log p + \log k\log(p-k))$. Compared with previous proximal operators for squared $k$-support norm, this complexity is better than that in~\cite{afs12}, and roughly the same as the most recent one in~\cite{mps14}.

\begin{algorithm}
\renewcommand{\algorithmicrequire}{\textbf{Input:}}
\renewcommand{\algorithmicensure} {\textbf{Output:} }\vspace{-1mm}
\caption{Algorithm for computing $\mathbf{prox}_{\mathbbm{I}_{\cC_{\lambda}}}(\cdot)$ of $\| \cdot \|_{k}^{sp^*}$}
\label{AlgProx}
\begin{algorithmic}[1]
\REQUIRE ~$\x$, $k$, $\lambda$ \\
\ENSURE ~$\w^* = \mathbf{prox}_{\mathbbm{I}_{\cC_{\lambda}}}(\x)$ \vspace{-1mm}\\
\IF {$\| \x \|_{k}^{sp^*} \leq \lambda$}
\STATE $\w^* := \x$
\ELSE
\STATE $l := 0$, $u := k-1$, and sort $| \x |$ to get $| \x |^{\downarrow}$
\WHILE {$l \leq u$}
\STATE $\tilde{s} := \lfloor (l+u) / 2 \rfloor$, and binary search for $\tilde{r}$ that satisfies \eqref{eq:cond2} and compute $\tilde{\w}$ based on \eqref{eq:projection}
\IF {$\tilde{r}$ does not exist}
\STATE $l := \tilde{s}+1$
\ELSIF {$\tilde{r}$ exists and \eqref{eq:cond1} is satisfied}
\STATE $\w^* := \tilde{\w}$, $l := \tilde{s}+1$
\ELSIF {$\tilde{r}$ exists but \eqref{eq:cond1} is not satisfied}
\STATE $u := \tilde{s}-1$
\ENDIF
\ENDWHILE
\ENDIF \vspace{-1mm}
\end{algorithmic}
\end{algorithm}

\subsection{Statistical Recovery Guarantees for $k$-support norm}
The analysis of the generalized Dantzig Selector for $k$-support norm consists of addressing two key challenges. First, note that Theorem~\ref{theo:dantzig recovery} requires an appropriate choice of $\lambda_p$. Second, one needs to compute the Gaussian width of the subset of the error set $\cT_\cR(\btheta^*)\cap \s^{p-1}$. For the $k$-support norm, we can get upper bounds to both of these quantities. We start by defining some notations. Let $\cG^* \subseteq \cG^{(k)}$ be the set of groups intersecting with the support of $\btheta^*$, and let $S$ be the union of groups in $\cG^*$, such that $s = |S|$. Then, we have the following bounds which are used for choosing $\lambda_p$, and bounding the Gaussian width.
\begin{theo}
For the $k$-support norm Generalized Dantzig Selection problem~\eqref{eq:ksup minimization}, we obtain
For the $k$-support norm Generalized Dantzig Selection problem~\eqref{eq:ksup minimization}, we obtain
\begin{align}
\E\left[ \cR^*(\bX^T \w)\right] &\leq   \sqrt{n} \left( \sqrt{2k\log\left(\frac{pe}{k}\right)} + \sqrt{k} \right) \label{eq:ksup lambda_p} \\
\omega(\Omega_\cR) &\leq \left( \sqrt{2k\log\left(\frac{pe}{k}\right)} + \sqrt{k} \right) \\
\omega(\cT_\cA(\btheta^*)\cap \s^{p-1})^2 &\leq \left(\sqrt{2k\log \left(p - k - \ceil{\frac{s}{k}} +2\right)}  + \sqrt{k}\right)^2 \cdot \ceil{\frac{s}{k}} + s\label{eq:ksup width}~.
\end{align}
\label{theo:ksup width}
\end{theo}
%
%
We prove these two bounds using the analysis technique for group lasso with overlaps developed in~\cite{rarn12}. Thereafter, choosing 
$\lambda_p =  \sqrt{n} \left( \sqrt{2k\log\left(\frac{pe}{k}\right)} + \sqrt{k} \right) $, 
and under the assumptions of Theorem~\ref{theo:dantzig recovery}, we obtain  the following result on the error bound for the minimizer of~\eqref{eq:ksup minimization}.

\begin{corr}
Suppose that all conditions of Theorem~\ref{theo:dantzig recovery} hold, and we solve~\eqref{eq:ksup minimization} with $\lambda_p$ chosen as above. Then, with high probability, we obtain
\begin{align}
\|\hat{\btheta}- \btheta^*\|_2 \leq \frac{4c \Psi_\cR \left( \sqrt{2k\log\left(\frac{pe}{k}\right)} + \sqrt{k} \right)   }{\kappa_\cL \sqrt{n}}
\end{align}
\label{corr:ksup recovery}
\end{corr}
%
{\bf Remark} The error bound provides a natural interpretation for the two special cases of the $k$-support norm, viz. $k=1$ and $k=p$. First, for $k=1$ the $k$-support norm is exactly the same as the $L_1$ norm, and the error bound obtained will be O$\left(\sqrt{\frac{s\log p}{n} }\right)$, the same as known results of DS, and shown in Section 2.2. Second, for $k=p$, the $k$-support norm is equal to the $L_2$ norm, and the error cone~\eqref{equ:tan cone} is then simply a half space (there is no structural constraint). Therefore, $\Psi_\cR = O(1)$, and the error bound scales as O$\left(\sqrt{\frac{p}{n}}\right)$.

%


\section{Experimental Results}
\label{sec:expt}
On optimization side, our ADMM framework is concentrated on its generality, and its efficiency has been shown in~\cite{wy12} for the special case of $L_1$ norm. Hence we focus on the efficiency of different proximal operators related to $k$-support norm. On statistical side, we concentrate on the behavior and performance of GDS with $k$-support norm. All experiments are implemented in MATLAB.

\subsection{Efficiency of Proximal Operator}
\begin{figure} [!hbt]
\centering
\includegraphics[width=6.5in]{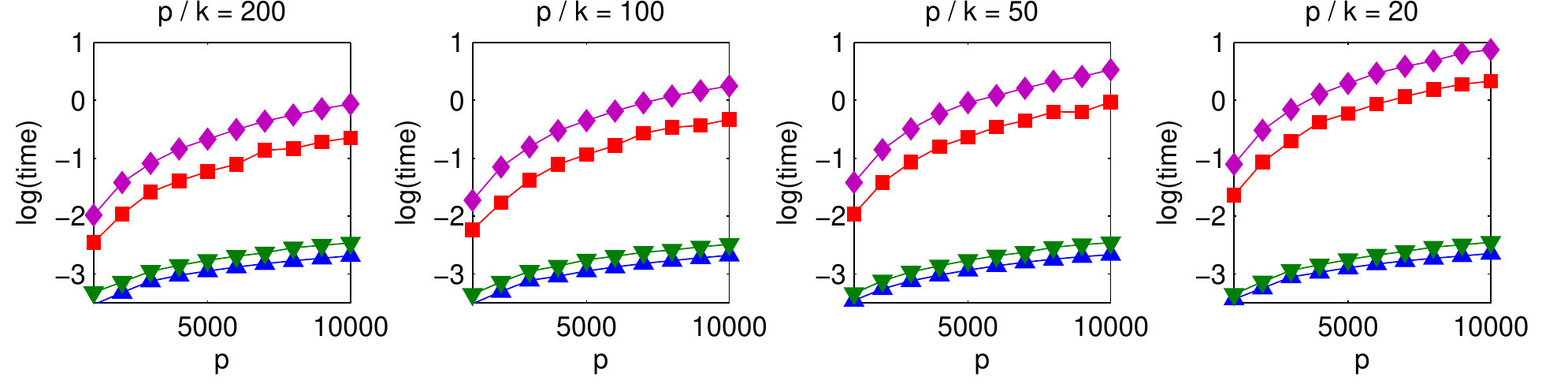}
\caption{Efficiency of proximal operators. Diamond: normal $\mathbf{prox}_{\mathbb{I}_{\cC_{\lambda}}}(\cdot)$, Square: $\mathbf{prox}_{\frac{1}{2 \beta} (\| \cdot \|_k^{sp})^2}(\cdot)$, Downward-pointing triangle: $\mathbf{prox}_{\frac{\lambda}{2} \| \cdot \|_{\Theta}^2}(\cdot)$, Upward-pointing triangle: accelerated $\mathbf{prox}_{\mathbb{I}_{\cC_{\lambda}}}(\cdot)$. For each ($p$, $k$), 200 vectors are randomly generated for testing.
\vspace{-5mm}}
\label{opti_exp}
\end{figure}

We tested four proximal operators related to $k$-support norm, which are our normal $\mathbf{prox}_{\mathbbm{I}_{\cC_{\lambda}}}(\cdot)$ and its accelerated version, $\mathbf{prox}_{\frac{1}{2 \beta} (\| \cdot \|_k^{sp})^2}(\cdot)$ in~\cite{afs12}, and $\mathbf{prox}_{\frac{\lambda}{2} \| \cdot \|_{\Theta}^2}(\cdot)$ in~\cite{mps14}. The dimension $p$ of vector in experiment varied from 1000 to 10000, and the ratio $p/k = \{200, 100, 50, 20\}$. As illustrated in Figure \ref{opti_exp}, in general, the speedup of accelerated $\mathbf{prox}_{\mathbbm{I}_{\cC_{\lambda}}}(\cdot)$ is considerable when compared with the normal $\mathbf{prox}_{\mathbbm{I}_{\cC_{\lambda}}}(\cdot)$ and $\mathbf{prox}_{\frac{1}{2 \beta} (\| \cdot \|_k^{sp})^2}(\cdot)$. Empirically it is also slightly better than the $\mathbf{prox}_{\frac{\lambda}{2} \| \cdot \|_{\Theta}^2}(\cdot)$.

\subsection{Statistical Recovery on Synthetic Data}

\textbf{Data generation} ~We fixed $p = 600$, and $\btheta^* = (\underbrace{10, \ldots, 10}_{10}, \underbrace{10, \ldots, 10}_{10}, \underbrace{10, \ldots, 10}_{10}, \underbrace{0, 0, \ldots, 0}_{570})$ throughout the experiment, in which nonzero entries were divided equally into three groups. The design matrix $\bX$ were generated from a normal distribution such that the entries in the same group have the same mean sampled from $\mathcal{N}(0,1)$. $\bX$ was normalized afterwards. The response vector $\y$ was given by $\y = \bX \btheta^* + 0.01 \times \mathcal{N}(0, 1)$. The number of samples $n$ is specified later.

\begin{figure*}[!h]
\centering

\subfigure[ROC curves] {
    \label{fig:roc}
    \includegraphics[width=0.33\linewidth]{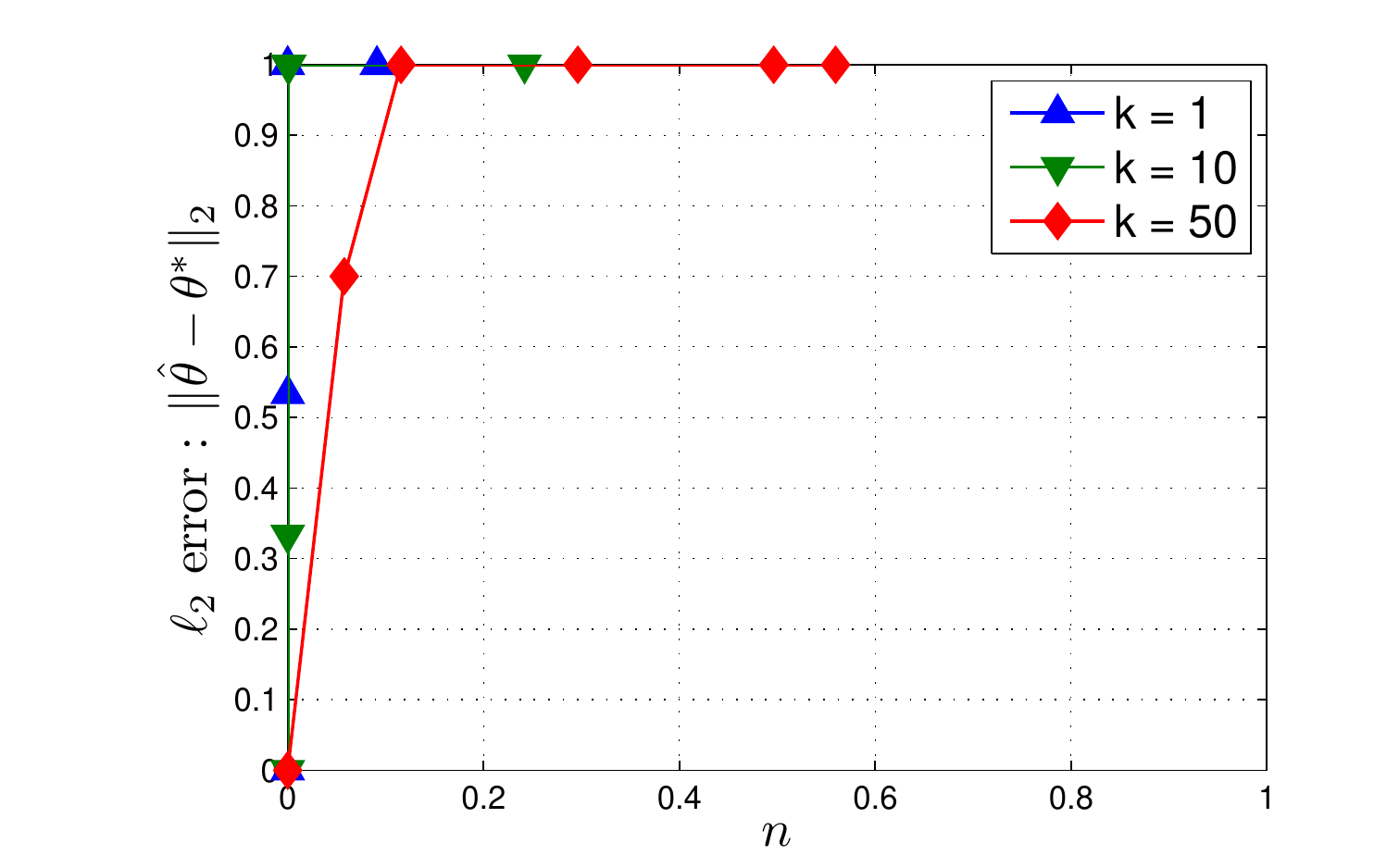}}
\begin{minipage}[b]{0.2in}

\end{minipage}
\subfigure[$L_2$ error vs. $n$] {
    \label{fig:l2error_n}
    \includegraphics[width=0.31\linewidth]{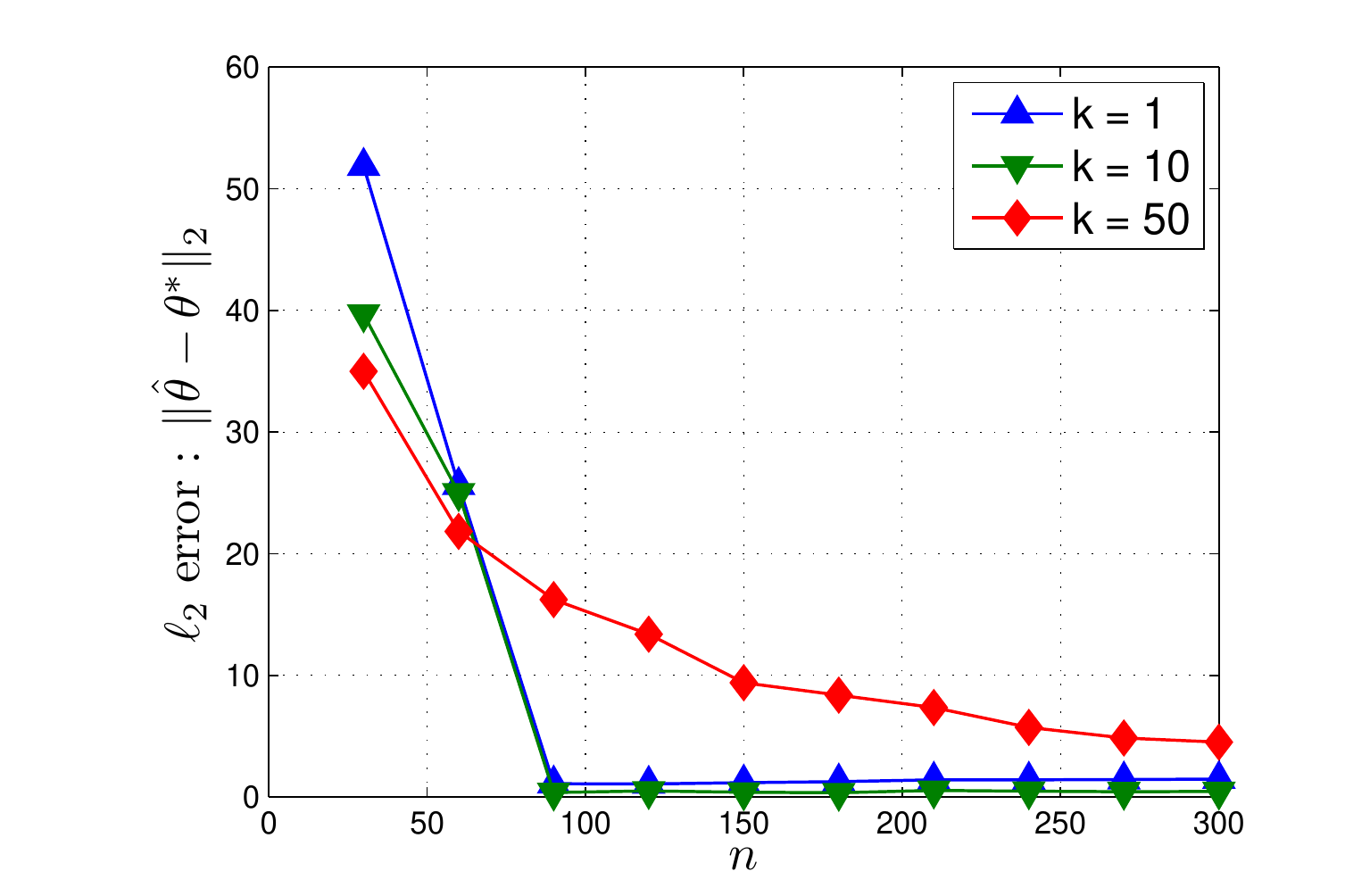}}
\begin{minipage}[b]{0.2in}

\end{minipage}
\subfigure[$L_2$ error vs. $k$] {
    \label{fig:l2error_k}
    \includegraphics[width=0.31\linewidth]{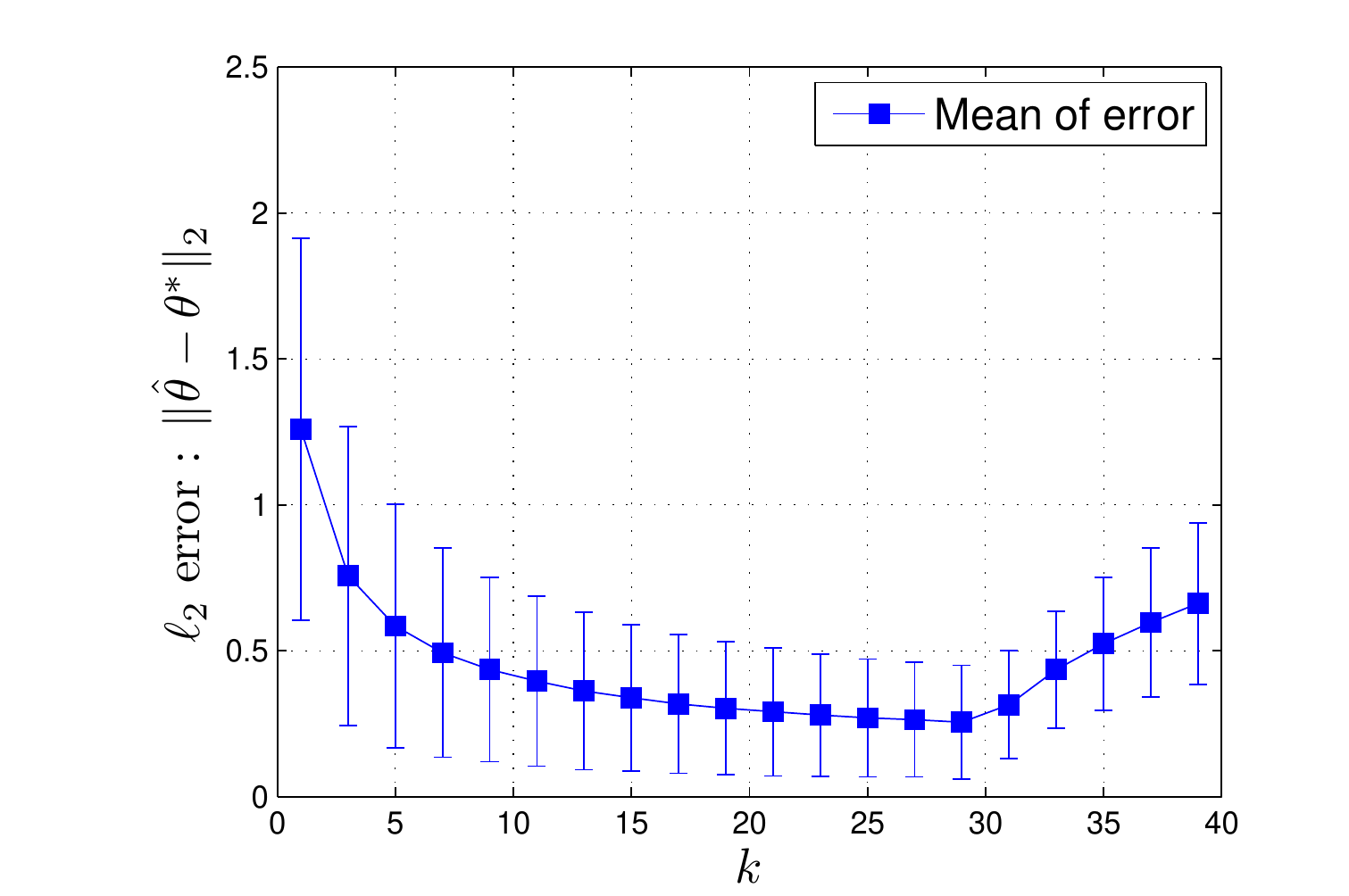}}

\caption{(a) The true positive rate reaches 1 quite early for $k = 1, 10$. When $k = 50$, the ROC gets worse due to the strong smoothing effect introduced by large $k$. (b) For each $k$, the $L_2$ error is large when the sample is inadequate. As $n$ increases, the error decreases dramatically for $k = 1,10$ and becomes stable afterwards, while the decrease is not that significant for $k = 50$ and the error remains relatively large. (c) Both mean and standard deviation of $L_2$ error are decreasing as $k$ increases until it exceeds the number of nonzero entries in $\btheta^*$, and then the error goes up for larger $k$, which matches our analysis quite well. The result also shows that the $k$-support-norm GDS with suitable $k$ outperforms the $L_1$ DS when correlated variables present in data (Note that $k=1$ corresponds to standard DS). \vspace{-3mm}}
\label{exp_stat}
\end{figure*}

\textbf{ROC curves with different $\boldsymbol k$} ~We fixed $n = 400$ to obtain the ROC plot for $k = \{1, 10, 50 \}$ as shown in Figure~\ref{fig:roc}. $\lambda_p$ ranged from $10^{-2}$ to $10^3$.


\textbf{$L_2$ error vs. $\boldsymbol n$} ~We investigated how the $L_2$ error $\| \htheta - \btheta^* \|_2$ of Dantzig selector changes as the number of samples increases, where $k = \{1, 10, 50 \}$ and $n = \{30, 60, 90, \ldots, 300\}$. The plot is shown in Figure~\ref{fig:l2error_n}.

\textbf{$L_2$ error vs. $\boldsymbol k$} ~We also looked at the $L_2$ error with different $k$. We again fixed $n = 400$ and varied $k$ from 1 to 39. For each $k$, we repeated the experiment 100 times, and obtained the mean and standard deviation plot in Figure~\ref{fig:l2error_k}.

\section{Conclusions}
\label{sec:conc}
In this paper, we introduced the GDS, which generalizes the standard $L_1$-norm Dantzig Selector to estimation with any norm, such that structural information encoded in the norm can be efficiently exploited. A flexible framework based on inexact ADMM is proposed for solving the GDS, which only requires one of conjugate proximal operators to be efficiently solved. Further, we provide a unified statistical analysis framework for the GDS, which utilizes Gaussian widths of certain restricted sets for proving consistency. In the non-trivial example of $k$-support norm, we showed that the proximal operators used in the inexact ADMM can be computed more efficiently compared to previously proposed variants. Our statistical analysis for the $k$-support norm provides the first result of consistency of this structured norm. Last, experimental results provided sound support to the theoretical development in the paper.

\section*{Acknowledgements}
The research was supported by NSF grant IIS-1029711. The work was also supported by NSF grants IIS- 0916750, IIS-0953274 and CNS-1314560 and by NASA grant NNX12AQ39A. A. B. acknowledges support from IBM and Yahoo.

\bibliographystyle{plain}
\bibliography{dantzsel_ref,mod_ref,supplement_ref}

\appendix

\section{Proof of Theorem~\ref{theo:dantzig recovery}}
{\bf Statement of Theorem:}\textit{
Suppose the design matrix $\bX$ consists of i.i.d. Gaussian entries with zero mean variance 1, and we solve the optimization problem~\eqref{eq:dantzig minimization} with
\beq
\lambda_p \geq c\E\left[ \cR^*(\bX^T\w)\right]~.
\label{eq:lambda choice}
\eeq
Then, with probability at least $(1 - \eta_1 \exp(-\eta_2 n))$, we have
\beq
\|\hat{\btheta} - \btheta^*\|_2 \leq \frac{4c \Psi_\cR \omega(\Omega_\cR)  }{\kappa_\cL \sqrt{n}}~,
\eeq
where $\omega(\cT_\cR(\btheta^*)\cap \s^{p-1})$ is the Gaussian width of the intersection of $\cT_\cR(\btheta^*)$ and the unit spherical shell $\s^{p-1}$, $\omega(\Omega_\cR)$ is the Gaussian width of the unit norm ball,  $\kappa_\cL>0$ is the gain given by
\beq
\kappa_\cL= \frac{1}{n}\left(\ell_n - \omega(\cT_\cR(\btheta^*)\cap \s^{p-1})\right)^2~, 
\eeq
$\Psi_\cR = \sup_{\Delta\in \cT_\cR}\cR(\Delta)/\|\Delta\|_2 $ is a norm compatibility factor,
$\ell_n$ is the expected length of a length $n$ i.i.d. standard Gaussian vector with $\frac{n}{\sqrt{n+1}} < \ell_n < \sqrt{n}$, and $c>1,\eta_1,\eta_2 >0$ are constants.
}

\proof
We use the following lemma for the proof.
\begin{lemm}
Suppose we solve the minimization problem~\eqref{eq:dantzig minimization} with $\lambda_p \geq \cR^*\left(\bX^T \w\right)$. Then the error vector $\hat{\Delta}$ belongs to the set
\beq
\cT_\cR(\btheta^*) := \text{\upshape cone}\left\{ \Delta \in \R^p~:~\cR(\btheta^* + \Delta) \leq \cR(\btheta^*)\right\}~,
\label{eq:tan cone}
\eeq
and the error $\hat{\Delta} = \hat{\btheta} - \btheta^*$ satisfies the following bound
\beq
\cR^*\left(\bX^T\bX \hat{\Delta}\right) \leq 2 \lambda_p
\eeq
\label{Mlem:error set}
\end{lemm}
\proof
By our choice of $\lambda_p$, both $\btheta^*$ and $\htheta$ lie in the feasible set of~\eqref{eq:dantzig minimization} , and by optimality of $\htheta$,
\beq
\cR\left(\btheta^* + \hDelta\right)  = \cR(\htheta) \leq \cR(\btheta^*)~.
\eeq
Also, by triangle inequality
\begin{align}
\cR^*\left(\bX^T\bX \hat{\Delta}\right) &= \cR^*\left(\bX^T\bX (\htheta - \btheta^*) \right) \\
&\leq \cR^*\left(\bX^T(\y - \bX\btheta^*) \right) +  \cR^*\left(\bX^T(\y - \bX\htheta) \right) \leq 2 \lambda_p~.
\end{align}
\qed

Now, note that $\bX$ and $\w$ are independent and we can rewrite
\beq
\E_{\bX,\w}\left[ \cR^*(\bX^T\w)\right] = \E_\w\left[ \E_\bX\left[ \cR^*(\bX^T\w)|\w\right]\right] = \E_\w\Bigg[\|\w\|_2 \E_\bX\left[ \cR^*\left(\bX^T\frac{\w}{\|\w\|_2}\right)|\w\right]\Bigg]~.
\eeq
Since $\w/\|\w\|_2$ is an isotropic unit vector uniformly distributed over the surface of the unit sphere, $\left(\bX^T\frac{\w}{\|\w\|_2}\right) = \g$ is an i.i.d. $\cN(0,1)$ Gaussian vector. Therefore
\beq
\E_{\bX,\w}\left[ \cR^*(\bX^T\w)\right]  = \E_\w[\|\w\|_2] \E_\g[\cR^*(\g)]~.
\eeq
Also, note that $\cR^*(\cdot)$ is Lipschitz continuous with Lipschitz constant of 1 w.r.t. the norm $\cR^*$, and hence by Gaussian concentration of Lipschitz functions~\cite{leta91}, 
\beq
\P\left( \cR^*(\g) \geq \E_\g[\cR^*(\g)] + \tau \right) \leq\exp\left[ -\frac{\tau^2}{2}\right]  ~,
\eeq
and similarly $\|\w\|_2 \leq \ell_n + \tau$ with probability at least $1 - \exp(-\tau^2/2)$, where $\frac{n}{\sqrt{n+1 }}\leq \ell_n \leq \sqrt{n}$ is the expected length of $\w$.
Therefore, for some $c >1$ choosing $\lambda_p \geq c\E\left[ \cR^*(\bX^T\w)\right] = c~\ell_n \E_\g[\cR^*(\g)]$ implies that
\beq
\P\left(\lambda_p \geq \cR^*(\bX^T\w)\right) \geq \left(1 - \exp\left[-\frac{c_1\E_\g^2[\cR^*(\g)]}{2}\right]\right) \left( 1 - \exp\left[- \frac{c_2\ell_n^2}{2}\right]\right) = 1 - \eta_1'\exp(-\eta_2'n)~,
\eeq
for some constant $c_1, c_2,\eta_1', \eta_2' >0$. Further, note that $\E_\g[\cR^*(\g)] = \omega(\Omega_\cR)$, the Gaussian width of the unit ball of norm $\cR$.

Also, from Lemma~\ref{Mlem:error set}, we have
\beq
\cR^*\left(\bX^T\bX \hat{\Delta}\right) \leq 2 \lambda_p
\eeq
Now, note that
\beq
\|\bX\hDelta\|_2^2 = \langle\hDelta, \bX^T \bX \hDelta\rangle \leq |\langle\hDelta, \bX^T \bX \hDelta\rangle| \leq   \cR(\hDelta) \cR^*\left(\bX^T \bX\hDelta \right) \leq 2 \lambda_p\cR(\hDelta) ~,
\label{eq:up bound gain}
\eeq
where we have used H\"{o}lder's inequality, and the bound $\cR^*\left(\bX^T \bX \hDelta\right) \leq 2\lambda_p$ from above.



Next, we use Gordon's theorem, which states that for $\bX$ with i.i.d. Gaussian $(0,1)$ entries,
\beq
\E\left[ \min_{\z \in \cT_\cR(\btheta^*)\cap \s^{p-1}} \|\bX \z\|_2 \right] \geq \ell_n - \omega\left(\cT_\cR(\btheta^*)\cap \s^{p-1}\right)~,
\eeq
where $\ell_n$ is the expected length of an i.i.d. Gaussian random vector of length $n$, and $\omega\left(\cT_\cR(\btheta^*)\cap \s^{p-1}\right)$ is the Gaussian width of the set $\Omega = \left(\cT_\cR(\btheta^*)\cap \s^{p-1}\right)$. Now, since the function $\bX \rightarrow \min_{\z \in \Omega}\|\bX\z\|_2$ is Lipschitz continuous with constant $1$ over the set $\Omega$, we can use Gaussian concentration of Lipschitz functions~\cite{leta91} to obtain
\begin{align}
&\|\bX\Delta\|_2 \geq \frac{1}{2} \left(\ell_n - \omega(\cT_\cR(\btheta^*)\cap \s^{p-1})\right) \|\Delta\|_2 \\
\Rightarrow ~~&\frac{1}{\sqrt{n}} \|\bX\Delta\|_2 \geq \frac{\left(\ell_n - \omega(\cT_\cR(\btheta^*)\cap \s^{p-1})\right) }{2\sqrt{n}} \|\Delta\|_2 \\
\Rightarrow ~~&\label{eq:gain cond}  \frac{1}{n} \|\bX\Delta\|_2^2 \geq \frac{\kappa_\cL}{2} \|\Delta\|_2^2 ~,
\end{align}
with probability greater than $1 - \exp\left(-\frac{1}{8} \left(\ell_n - \omega(\cT_\cR(\btheta^*)\cap \s^{p-1})\right)^2\right) = 1- \eta_1''\exp(-\eta_2''n)$, where $\kappa_\cL= \left(\ell_n - \omega(\cT_\cR(\btheta^*)\cap \s^{p-1})\right)^2 /n >0$ is the \emph{gain}, and $\eta_1'', \eta_2''>0$ are constants .

%
Combining~\eqref{eq:gain cond} and \eqref{eq:up bound gain}, and using the choice of $\lambda_p$, we obtain

\beq
\|\hat{\btheta}_n - \btheta^*\|_2  = \|\hDelta\|_2 \leq \frac{4c \E\left[ \cR^*(\bX^T\w)\right]  }{\kappa_\cL n} \frac{\cR(\Delta)}{\|\Delta\|_2} \leq \frac{4c\Psi_\cR \omega(\Omega_\cR)  }{\kappa_\cL \sqrt{n}} 
\eeq
with probability greater than $(1 - \eta_1'\exp(-\eta_2'n)) (1 - \eta_1''\exp(-\eta_2''n)) = 1 - \eta_1\exp(-\eta_2n)$, for constants $\eta_1, \eta_2$ where
\beq
\Psi_\cR = \sup_{\Delta \in \cT_\cR}\frac{\cR(\Delta)}{\|\Delta\|_2}~.
\eeq
The statement of the theorem follows.
\qed

\section{Proof of Theorem~\ref{maintheorem}}
{\bf Statement of Theorem:}\textit{
Given $\lambda > 0$ and $\x \in \R^p$, if $\|\x\|_{k}^{sp^*} \leq \lambda$, then $\w^*=\mathbf{prox}_{\mathbbm{I}_{\cC_{\lambda}}}(\x)=\x$. If
$\|\x\|_{k}^{sp^*} > \lambda$, define $A_{sr} = \sum_{i=s+1}^r |\x|_i^{\downarrow}$, $B_{s} = \sum_{i=1}^{s}(|\x|_i^{\downarrow})^2$, in which
$0 \leq s < k$ and $k \leq r \leq p$, and construct the nonlinear equation of $\beta$,
\begin{equation}
\label{nonlinearequ}
(k-s)A_{sr}^2 \left[\frac{1+\beta}{r-s+(k-s)\beta} \right]^2 - {\lambda}^2(1+\beta)^2 + B_s = 0 ~.
\end{equation}
Let $\beta_{sr}$ be given by
\begin{equation}
\beta_{sr} = \left \{
             \begin{array}{lll}
              \text{nonnegative root of \eqref{nonlinearequ}} \ &\text{if $s > 0$ and the root exists} \\
              0 \ &\text{otherwise}
             \end{array}  \right. ~.
\end{equation}
Then the proximal operator $\w^*=\mathbf{prox}_{\I_{ \cC_{\lambda}}}(\x)$ is given by
\begin{equation}
\label{projection}
|\w^*|_i^{\downarrow} = \left \{
             \begin{array}{lll}
             \frac{1}{1+\beta_{s^*r^*}} |\x|_i^{\downarrow} \ &\text{if} \ \ 1 \leq i \leq s^*   \\
             \sqrt{\frac{{\lambda}^2 - B_{s^*}}{k-s^*}} \ &\text{if} \ \ s^* < i \leq r^* \ \text{and} \ \beta_{s^*r^*} = 0 \\
             \frac{A_{s^*r^*}}{r^*-s^*+(k-s^*)\beta_{s^*r^*}} \ &\text{if} \ \ s^* < i \leq r^* \ \text{and} \ \beta_{s^*r^*} > 0 \\
             |\x|_i^{\downarrow} \ &\text{if} \ \ r^* < i \leq p
             \end{array}  \right. ~,
\end{equation}
where the indices $s^*$ and $r^*$ with computed $|\w^*|^{\downarrow}$ make the following two inequalities hold,
\begin{equation}
\label{cond1}
|\w^*|_{s^*}^{\downarrow} > |\w^*|_{k}^{\downarrow} ~,
\end{equation}
\begin{equation}
\label{cond2}
|\x|_{r^* + 1}^{\downarrow} \leq |\w^*|_{k}^{\downarrow} < |\x|_{r^*}^{\downarrow} ~.
\end{equation}
There might be multiple pairs of $(s, r)$ satisfying the inequalities \myref{cond1}-\myref{cond2}, and we choose the pair with the smallest $\| |\x|^{\downarrow} - |\w|^{\downarrow} \|_2$. Finally, $\w^*$ is obtained by sign-changing and reordering $|\w^*|^{\downarrow}$ to conform to $\x$.
}

\proof Let $\w^* = \mathbf{prox}_{\mathbbm{I}_{\cC_{\lambda}}}(\x) = \argmin_{\w \in \cC_{\lambda}} \frac{1}{2} \|\x - \w\|_2^2$. For simplicity, we drop the constant $\frac{1}{2}$ in later discussion. Given a vector $\x$, we use the notation $\x_{i:j}$ to denote its subvector $(\x_i, \x_{i+1}, \ldots, \x_j)$. We consider the following two cases. \par
\textbf{Case 1:} if $\|\x\|_{k}^{sp^*} \leq \lambda$, it is trivial that $\w^* = \x$, which is also the global minimizer of $\|\x-\w\|_2^2$ without the constraint $\x \in \cC_{\lambda}$.

\textbf{Case 2:} if $\|\x\|_{k}^{sp^*}>\lambda$, first we start by noting that given $\x$ and $\w$, the following inequality holds
\begin{align*}
\|\x-\w\|_2^2 &= \|\x\|_2^2 - 2\langle \x,\w \rangle + \|\w\|_2^2 \\
&\geq \|\x\|_2^2 - 2 \langle |\x|^{\downarrow},|\w|^{\downarrow} \rangle + \|\w\|_2^2 ~,
\end{align*}
which implies that $\w^*$ should achieve this lower bound by conforming with the signs and orders of elements in $\x$. Without loss of generality, we are simply focused on the case where $\x = |\x|^{\downarrow}$.

For $\w^*$ to be the optimal, $\w_{k:p}^*$ should be chosen such that $\w^*_{k:r} = (\w_k^*, \w_k^*, \ldots, \w_k^*)$ and $\w^*_{r+1:p} = \x^*_{r+1:p}$, where $r$ satisfies
\begin{align*}
\x_{r} > \w^*_k \geq \x_{r+1} ~,
\end{align*}
otherwise either the decreasing order of $\w^*$ will be violated or the $\|\x_{k:p} - \w_{k:p} \|_2$ is not minimized. As for $\w^*_{1:k-1}$, we similarly assume $\w^*_{s+1:k-1} = (\w_k^*, \w_k^*, \ldots, \w_k^*)$ for some $0 \leq s \leq k-1$, then $\w^*_{1:s}$ should be chosen to minimize $\|\x_{1:s}-\w_{1:s}\|_2$ such that
\begin{align*}
\|\w_{1:s}\|_2^2 = \|\w_{1:k}^*\|_2^2 - \|\w_{s+1:k}^*\|_2^2 \leq {\lambda}^2 - (k-s)(\w^*_k)^2.
\end{align*}
By Cauchy-Schwarz Inequality, we have
\begin{align*}
\|\x_{1:s}-\w_{1:s}\|_2^2 \geq \|\x_{1:s}\|_2^2 - 2 \| \x_{1:s} \|_2 \| \w_{1:s} \|_2  + \|\w_{1:s}\|_2^2 ~,
\end{align*}
where the equality holds when $\w^*_{1:s}$ follows the form of $\w^*_{1:s} = \frac{1}{1+\beta_{sr}} \x_{1:s}$, and $\beta_{sr} \geq 0$ satisfies the constraint $\frac{B_s}{(1+\beta_{sr})^2} = {\lambda}^2 - (k-s)(\w_k)^2$.

So far we have figured out the structure of $\w^* = (\w^*_{1:s}, \w^*_{s+1:r}, \w^*_{r+1:p})$, in which the three subvectors, compared with $\x$, are shrunk by a common factor $1+\beta_{sr}$, constant $\w^*_k$, or unchanged. Next we need to determine the value of $\beta_{sr}$ and $\w^*_k$. By optimality, $\|\x-\w\|_2^2 = \|\x_{1:r}-\w_{1:r}\|_2^2$ must be minimized at $\w^*$, so we have the following problem,
\begin{equation}
\label{projobj}
\begin{split}
\min_{\beta, \w_k} \ \|\x_{1:r}-\w_{1:r}\|_2^2 &= \|\x_{1:s}-\w_{1:s}\|_2^2 + \|\x_{s+1:r}-\w_{s+1:r}\|_2^2 \\
&= (\frac{\beta}{1+\beta})^2 B_s + \sum_{i=s+1}^{r} (\x_i - \w_k)^2 \\
\end{split}
\end{equation}
\begin{equation}
\label{projcnstrt}
\text{s.t.} \ \ \ (\|\w\|_{k}^{sp^*})^2 = \frac{B_s}{(1+\beta)^2} + (k-s)(\w_k)^2 = {\lambda}^2
\end{equation}
Replacing $\w_k$ in \eqref{projobj} with $\w_k = \sqrt{\frac{{\lambda}^2 - \frac{B_s}{(1+\beta)^2}}{k-s}}$ obtained from \eqref{projcnstrt}, we express $\|\x_{1:r}-\w_{1:r}\|_2^2$ as a function of $\beta$,
\begin{equation}
\label{projobj_new}
\Phi_{sr}(\beta) = (\frac{\beta}{1+\beta})^2 B_s + \sum_{i=s+1}^{r} \big (\x_i - \sqrt{\frac{\lambda^2 - \frac{B_s}{(1+\beta)^2}}{k-s}} \big )^2
\end{equation}
Set derivative of $\Phi_{sr}(\beta)$ to be zero, we have
\begin{align}
\label{zeropoint1}
\frac{d}{d \beta} \Phi_{sr}(\beta) &= \frac{d}{d \beta} \Big [ (\frac{\beta}{1+\beta})^2 B_s + \sum_{i=s+1}^{r} \big (\x_i - \sqrt{\frac{\lambda^2 - \frac{B_s}{(1+\beta)^2}}{k-s}} \big )^2 \Big ] \\
\label{zeropoint2}
&= \frac{2\beta}{(1+\beta)^3}B_s - \frac{2A_{sr}B_s}{(1+\beta)^3(k-s)\sqrt{\frac{\lambda^2 - \frac{B_s}{(1+\beta)^2}}{k-s}}} +
\frac{2(r-s)B_s}{(k-s)(1+\beta)^3} \\
\label{zeropoint3}
&= \frac{2B_s}{(k-s)(1+\beta)^3} \Big [ (k-s)\beta - \frac{A_{sr}}{\sqrt{\frac{\lambda^2 - \frac{B_s}{(1+\beta)^2}}{k-s}}} + (r-s)\Big] = 0
\end{align}
If $s > 0$, then $B_s > 0$ and \eqref{zeropoint3} is equivalent to \eqref{nonlinearequ}. And we can see that the quantity inside the bracket of \eqref{zeropoint3} is monotonically increasing when $\beta \geq \max(0, \frac{\sqrt{B_s}- \lambda}{\lambda})$, thus ensuring the nonnegative root $\beta_{sr}$ is unique if it exists. If the nonnegative root exists, the expression for $\w^*_{s+1:r}$ can be obtained from \eqref{zeropoint3}, whose entries are all equal to $\w^*_k$.

If $s > 0$ and a nonnegative root of \eqref{zeropoint3} is nonexistent, the derivative is always positive when $\beta \geq 0$, which means that $\Phi_{sr}(\beta)$ is increasing. Hence the minimizer of $\Phi_{sr}(\beta)$ is $\beta_{sr} = 0$. If $s = 0$, we actually do not care about the value of $\beta_{sr}$ because the problem defined by \eqref{projobj} and \eqref{projcnstrt} is independent of $\beta$, and we set it to be 0 for simplicity. According to \eqref{projcnstrt}, both cases of $\beta_{sr} = 0$ lead to the same expression for $\w^*_{s+1:r}$ in \eqref{projection}.

As we do not know beforehand which $s$ and $r$ to choose, we need to search for $s^*$ and $r^*$ that give the smallest $\| |\x|^{\downarrow} - |\w|^{\downarrow} \|_2$, and also need to check whether the $\w^*$ obtained by \eqref{projection} is in decreasing order, which are the conditions \eqref{cond1} and \eqref{cond2} presented in Theorem~\ref{maintheorem}. \qed

\section{Proof of Theorem~\ref{bisection}}

To prove Theorem~\ref{bisection}, we first need the following corollary from Theorem~\ref{maintheorem}.

\begin{corr}
\label{monotonicity}
When $\beta \geq \max(0, \frac{\sqrt{B_s}- \lambda}{\lambda})$, $\Phi_{sr}(\beta)$ defined in \eqref{projobj_new}
is decreasing when $\beta < \beta_{sr} $, and increasing when $\beta > \beta_{sr}$. Equivalently, $\Phi_{sr}(\beta) = \|\x_{1:r}-\w_{1:r}\|_2^2$, when treated as function of $\w_k$, is decreasing when $\w_k < \w^*_k$ and increasing when $\w_k > \w^*_k$.
\end{corr}
\proof The first part simply follows the monotonicity of $\frac{d}{d \beta} \Phi_{sr}(\beta)$ mentioned in the proof of Theorem \ref{maintheorem}, which implies that $\frac{d}{d \beta} \Phi_{sr}(\beta)$ is negative when $\beta < \beta_{sr}$, and positive when $\beta > \beta_{sr}$ . The constraint \eqref{projcnstrt} implies that $\w_k$ increases as $\beta$ increases. So $\|\x_{1:r}-\w_{1:r}\|_2^2$, as a function of $\w_k$, has the same monotonicity w.r.t. $\w_k$.  \qed

{\bf Statement of Theorem:}\textit{
In search of ($s^*$, $r^*$) defined in Theorem \ref{maintheorem}, there can be only one $\tilde{r}$ for a given candidate $\tilde{s}$ of $s^*$, such that the inequality \eqref{cond2} is satisfied. Moreover if such $\tilde{r}$ exists, then for any $r < \tilde{r}$, the associated $|\tilde{\w}|^{\downarrow}_k$ violates the first part of \eqref{cond2}, and for $r > \tilde{r}$, $|\tilde{\w}|^{\downarrow}_k$ violates the second part of \eqref{cond2}. On the other hand, based on the $\tilde{r}$, we have following assertion of $s^*$,
\begin{equation}
\label{s_start}
s^* \left \{
             \begin{array}{lll}
             > \tilde{s} \ &\text{if $\tilde{r}$ does not exist}  \\
             \geq \tilde{s} \ &\text{if $\tilde{r}$ exists and corresponding $|\tilde{\w}|^{\downarrow}_k$ satisfies \eqref{cond1}} \\
             < \tilde{s} \ &\text{if $\tilde{r}$ exists but corresponding $|\tilde{\w}|^{\downarrow}_k$ violates \eqref{cond1}}
             \end{array} \right. ~.
\end{equation}
}

\proof We again focus on the case of $\x = |\x|^{\downarrow}$. First we show by contradiction that for a given $\tilde{s}$, the $\tilde{r}$ that satisfies \eqref{cond2} can be at most one.

Suppose there are two indices, say $r_1$ and $r_2$, which satisfy that condition with the same $\tilde{s}$. Without loss of generality, let $r_1 < r_2$, we know that their corresponding $\w^{(1)}$ and $\w^{(2)}$ should minimize $\|\x_{1:r_1}-\w_{1:r_1}\|_2^2$ and $\|\x_{1:r_2}-\w_{1:r_2}\|_2^2$, respectively. As $r_1 < r_2$, then $\w_{k}^{(1)} \geq \x_{r_2} > \w_{k}^{(2)}$ according to \eqref{cond2}. Construct \\
\\
\centerline{
$\w' = (\underbrace{\frac{\x_1}{1+\beta'}, \ldots, \frac{\x_{\tilde{s}}}{1+\beta'}}_{\tilde{s}}, \underbrace{\x_{r_2}, \ldots, \x_{r_2}}_{r_2 - \tilde{s}}, \x_{r_2+1}, \ldots, \x_p)$}
where $\beta'$ is chosen to satisfy the constraint \eqref{projcnstrt} with $\w'_k = \x_{r_2}$, and $\|\x_{1:r_2}-\w^{(2)}_{1:r_2}\|_2^2$
can be decomposed as
\begin{align*}
\|\x_{1:r_2}-\w^{(2)}_{1:r_2}\|_2^2 &= \|\x_{1:r_1}-\w^{(2)}_{1:r_1}\|_2^2 + \|\x_{r_1+1:r_2}-\w^{(2)}_{r_1+1:r_2}\|_2^2 \\
&> \|\x_{1:r_1}-\w'_{1:r_1}\|_2^2 + \|\x_{r_1+1:r_2}-\w'_{r_1+1:r_2}\|_2^2 \\
& = \|\x_{1:r_2}-\w'_{1:r_2}\|_2^2
\end{align*}
which contradicts that $\w^{(2)}_{1:r_2}$ minimizes $\|\x_{1:r_2}-\w_{1:r_2}\|_2^2$. Note that $\|\x_{1:r_1}-\w^{(2)}_{1:r_1}\|_2^2 >
\|\x_{1:r_1}-\w'_{1:r_1}\|_2^2$ simply follows Corollary \ref{monotonicity} as $\w_{k}^{(1)} \geq \x_{r_2} = \w'_k  > \w_{k}^{(2)}$, and $\|\x_{r_1+1:r_2}-\w^{(2)}_{r_1+1:r_2}\|_2^2 > \|\x_{r_1+1:r_2}-\w'_{r_1+1:r_2}\|_2^2$ is due to the fact that $\x_{r_1+1} \geq \ldots \geq  \x_{r_2} = \w'_k > \w_{k}^{(2)}$.

Next we show by contradiction that if $\tilde{r}$ exists for given $\tilde{s}$, then any $r < \tilde{r}$ violates the first part of
\eqref{cond2}, and any $r > \tilde{r}$ violates the second part.

Let $\tilde{\w}$ denote the minimizer of $\|\x_{1:\tilde{r}}-\w_{1:\tilde{r}}\|_2^2$. Suppose $r < \tilde{r}$ and the first part of \eqref{cond2} is not violated, then its second part must be violated due to the uniqueness of $\tilde{r}$. Then we can construct new
\begin{align*}
\w' = (\underbrace{\frac{\x_1}{1+\beta'}, \ldots, \frac{\x_{\tilde{s}}}{1+\beta'}}_{\tilde{s}}, \underbrace{\x_{\tilde{r}}, \ldots, \x_{\tilde{r}}}_{\tilde{r}-\tilde{s}}, \x_{\tilde{r}+1}, \ldots, \x_p) ~,
\end{align*}
where $\beta'$ is again chosen to satisfy the constraint \eqref{projcnstrt} with $\w'_k = \x_{\tilde{r}}$. This by the same argument for proving the uniqueness of $\tilde{r}$ make the following inequality hold,
\begin{align*}
\|\x_{1:\tilde{r}}-\tilde{\w}_{1:\tilde{r}}\|_2^2 &= \|\x_{1:r}-\tilde{\w}_{1:r}\|_2^2 + \|\x_{r+1:\tilde{r}}-\tilde{\w}_{r+1:\tilde{r}}\|_2^2 \\
&> \|\x_{1:r}-\w'_{1:r}\|_2^2 + \|\x_{r+1:\tilde{r}}-\w'_{r+1:\tilde{r}}\|_2^2 \\
&= \|\x_{1:\tilde{r}}-\w'_{1:\tilde{r}}\|_2^2 ~.
\end{align*}
This contradicts that $\tilde{\w}$ is the minimizer of $\|\x_{1:\tilde{r}}-\w_{1:\tilde{r}}\|_2^2$. Similar argument applies to the case when $r > \tilde{r}$. Let $\beta''$ satisfy \eqref{projcnstrt} together with $\w''_k = \x_{r+1}$, and we construct
\begin{align*}
\w'' = (\underbrace{\frac{\x_1}{1+\beta''}, \ldots, \frac{\x_s}{1+\beta''}}_{\tilde{s}}, \underbrace{\x_{r+1}, \ldots, \x_{r+1}}_{r-\tilde{s}}, \x_{r+1}, \ldots, \x_p) ~,
\end{align*}
which gives smaller $\|\x_{1:r}-\w_{1:r}\|_2^2$ than any $\w$ with $\w_k < \x_{r+1}$. Therefore it is impossible for $r > \tilde{r}$ to violate the first inequality.

Finally we show the assertion \eqref{s_start} for $s^*$.

We note that given $\tilde{s}$ , finding solution to the proximal operator can be viewed as minimization of \eqref{projobj} under the constraint $\|\w_{1:k}\|_2 \leq \lambda$ and $\w_k = \w_{k-1} = \ldots = \w_{\tilde{s}+1}$. So for $s < \tilde{s}$, the minimization problem is equivalent to the one for $\tilde{s}$ under additional constraint $\w_{\tilde{s}+1} = \w_{\tilde{s}} = \ldots = \w_{s+1}$. If the $\tilde{r}$ does not exist, for $s < \tilde{s}$, $\tilde{r}$ is nonexistent either, thus $s^* > \tilde{s}$. If the $\tilde{r}$ exists and \eqref{cond1} is satisfied, then $s^* \geq \tilde{s}$ because $s < \tilde{s}$ considers a more restricted problem and is unable to obtain a smaller $\| \x - \w \|_2$.

For the situation in which $\tilde{r}$ exists for $\tilde{s}$ but the associated $\tilde{\w}_k$ violates \eqref{cond1}, we show by contradiction that for any $s > \tilde{s}$, \eqref{cond1} is also violated. 

Assume that $\w'$ (different from the previously used) satisfies both \eqref{cond1} and \eqref{cond2} for $s' = \tilde{s} + 1$ and the corresponding $r'$. It is not difficult to see that $\w'_k < \tilde{\w}_k$ and $r' \geq \tilde{r}$, otherwise $\| \w'_{1:k} \|_2 > \lambda$. By the violation we have shown for $r$, the minimizer of \eqref{projobj} for ($s', \tilde{r}$), denoted by $\w''$, satisfies $\w''_k \leq \w'_k$ (Note that $\w'$ is the minimizer of \eqref{projobj} for ($s', r'$) and $r' \geq \tilde{r}$). Combined with $\w'_k < \tilde{\w}_k$, this indicates by Corollary \ref{monotonicity} that $\Phi_{s' \tilde{r}}(\cdot)$ is increasing on the interval [$\w''_k, \tilde{\w}_k$]. Then we consider two sequential modifications on $\tilde{\w}$,
\begin{enumerate}
\item Replacing the $\tilde{\w}_{1:s'}$ in $\tilde{\w}$ with $\frac{\| \tilde{\w}_{1:s'} \|_2}{\| \x_{1:s'} \|_2} \x_{1:s'}$ ~,
\item Decreasing $\tilde{\w}_{s'+1:\tilde{r}}$ by certain amount and amplifying the new $\tilde{\w}_{1:s'}$ by some factor, such that \eqref{projcnstrt} still holds for $s'$ and $\tilde{\w}_{s'+1} = \tilde{\w}_{s'}$ ~.
\end{enumerate}
Note that the two modifications both decrease $\|\x_{1:\tilde{r}} - \tilde{\w}_{1:\tilde{r}} \|_2$. Decrease in Modification 1 is the result of Cauchy Schwarz Inequality, and decrease in Modification 2 is due to the monotonicity of $\Phi_{s' \tilde{r}}(\cdot)$ we mentioned afront. The modified $\tilde{\w}$ satisfies $\tilde{\w}_{\tilde{s}+1} = \tilde{\w}_{\tilde{s}+2} = \ldots = \tilde{\w}_{k}$, thus contradicting that the old $\tilde{\w}$ is the minimizer of \eqref{projobj} for ($\tilde{s}, \tilde{r}$). Hence, by induction, we conclude that for any $s' > \tilde{s}$, its solution also violates \eqref{cond1}.

Assembling the conclusions above, we have \eqref{s_start} for $s^*$. \qed

\section{Proof of Theorem~\ref{theo:ksup width}}
{\bf Statement of Theorem:}\textit{
For the $k$-support norm Generalized Dantzig Selection problem~\eqref{eq:ksup minimization}, we obtain
\begin{align}
\E\left[ \cR^*(\bX^T \w)\right] &\leq   \sqrt{n} \left( \sqrt{2k\log\left(\frac{pe}{k}\right)} + \sqrt{k} \right) \label{eq:ksup lambda_p} \\
\omega(\Omega_\cR) &\leq \left( \sqrt{2k\log\left(\frac{pe}{k}\right)} + \sqrt{k} \right) \\
\omega(\cT_\cA(\btheta^*)\cap \s^{p-1})^2 &\leq \left(\sqrt{2k\log \left(p - k - \ceil{\frac{s}{k}} +2\right)}  + \sqrt{k}\right)^2 \cdot \ceil{\frac{s}{k}} + s\label{eq:ksup width}~.
\end{align}
}

\proof
We first illustrate that the $k$-support norm is an atomic norm, and then prove Theorem~\ref{theo:ksup width}.
\subsection{$k$-Support norm as an Atomic Norm}
Here we show that $k$-support norm satisfies the definition of atomic norms~\cite{crpw12}.
Consider $\cG_j$ to be the set of all subsets of $\{1,2,\ldots,p\}$ of size $j$, so that
\beq
\cG^{(k)} = \left\{ \cG_j\right\}_{j=1}^k~.
\eeq
For every $j$, consider the set
\beq
\cA_j = \{ \w~:~\|(\w_{G_j}) \|_2 =1,~G_j \in \cG_j,~ \w_i=\frac{1}{\sqrt{j}},~~\forall i \in G_j,~\w_i =0, \forall i \notin G_j\}~,
\eeq
corresponding to $\cG_j$, and the union of such sets
\beq
\cA = \{\cA_j\}_{j\in \{1,\ldots,k\}}~.
\eeq
Note that since every non-zero element in a vector in $\cA_j$ is $\frac{1}{\sqrt{j}}$, such an element cannot be represented as a convex combination of elements of the set $\cA_l,~l<j$, whose non-zero elements are $\frac{1}{\sqrt{l}}$. Therefore none of the elements $\w$ in the set $\cA$ lies in the convex hull of the other elements $\cA\setminus\{\w\}$. Further, note that
\beq
\text{\normalfont conv} (\cA) = C_k~,
\eeq
and the $k$-support norm defines the gauge function of the $\cA$. Thus the $k$-support norm is an atomic norm.

\subsection{The Error set and its Gaussian width}
Note that the cardinality of the set $\cG^{(k)}$ is
\beq
M = {p \choose k} + {p \choose k-1}+{p \choose k-2}+ \cdots+{p \choose 1}
\eeq
The error set is given by
\beq
\cT_\cA(\btheta^*) = \text{cone}\{\Delta \in \R^p~:~\ksup{\Delta + \btheta^*} \leq \ksup{\btheta^*}\}~.
\label{eq:ksup_tancone}
\eeq
Note that this set is a cone, and we can define the \emph{normal} cone of this set as
\begin{align}
\cN_\cA(\btheta^*) &= \{\u~:~\langle\u,\Delta\rangle \leq 0,~\forall \Delta\in\cT_\cA(\btheta^*)\} \\
\end{align}
The following proposition, shown in~\cite{rarn12}, shows that the normal cone can be written in terms of the dual norm of the $k$-support norm.
\begin{prop}
The normal cone to the tangent cone defined in~\eqref{eq:ksup_tancone} can written as
\beq
\label{eq:normal cone} \cN_\cA(\btheta^*) = \{\u~:~\exists t>0 \text{ s.t. } \langle\u,\btheta^*\rangle=t\ksup{\btheta^*},~\ksupd{\u} \leq t\}~.
\eeq
\end{prop}
We provide a simple proof of this statement for our case for ease of understanding.

\proof
We re-write the definition of the normal cone in terms of the estimated parameter $\hat{\btheta}$ as
\beq
\cN_\cA(\btheta^*) = \{\u \in \R^p~:~\langle\u,\btheta - \btheta^*\rangle \leq 0, \forall \btheta - \btheta^* \in \cT_\cA(\btheta^*)\}~.
\eeq
Note that this means that $\u \in \cN_\cA(\btheta^*)$ if and only if
\begin{align}
&\langle\u, \btheta - \btheta^*\rangle \leq 0,~~ \forall \ksup{\btheta} \leq \ksup{\btheta^*}\\
\Rightarrow &\langle\u, \btheta\rangle \leq \langle\u, \btheta^*\rangle~~\forall \ksup{\btheta} \leq \ksup{\btheta^*}~.
\end{align}
Now, we claim that $\langle\u, \btheta^*\rangle\geq 0$ for all such $\u$. This can be shown as follows. Assume the contrary, i.e. there exists a $\hat{\u}\in \cN_\cA(\btheta^*)$ such that $\langle\hat{\u}, \btheta^*\rangle< 0$. Now, noting that $(-\btheta^*) \in \cT_\cA(\btheta^*)$, we have
\beq
\langle\hat{\u}, - \btheta^* \rangle = - \langle\hat{\u},\btheta^* \rangle  >0~,
\eeq
so that $\hat{\u} \notin \cN_\cA(\btheta^*)$, which is a contradiction, and the claim follows.

Therefore, we can write
\beq
\langle\u,  \btheta^* \rangle =t\ksup{\btheta^*}
\eeq
for some $t\geq 0$. Then, $\u \in \cN_\cA(\btheta^*)$ if and only if
\begin{align}
\exists t\geq 0~,~ \langle\u,  \btheta^* \rangle  = t\ksup{\btheta^*}~~,~~\langle\u, \btheta\rangle \leq t \ksup{\btheta^*}~~\forall \ksup{\btheta} \leq \ksup{\btheta^*}~.
\end{align}
Since
\beq
\langle\u, \btheta\rangle \leq t \ksup{\btheta^*},~~\forall \ksup{\btheta} \leq \ksup{\btheta^*} ~\Rightarrow~ \ksupd{\u} \leq t~,
\eeq
the statement  follows.
\qed

The $k$-support norm can be thought of as a group sparse norm with overlaps, such as been dealt with in~\cite{rarn12}. Therefore, we can utilize some of the analysis techniques developed in~\cite{rarn12}, specialized to the structure of the $k$-support norm. We begin by stating a theorem which enables us to bound the Gaussian width of the error set. Henceforth, we write $\cN_\cA = \cN_\cA(\btheta^*)$ and $\cT_\cA = \cT_\cA(\btheta^*)$ where the dependence on $\btheta^*$ is understood.

First, we define sets that involve the support set of $\btheta^*$. Let us define the set $\cG^* \subseteq \cG^{(k)}$ to be the set of all groups in $\cG^{(k)}$ which overlap with the support of $\btheta^*$, i.e.
\beq
\cG^* = \{G\in \cG^{(k)}~:~ G \cap \text{supp}(\btheta^*) \neq \emptyset \}~.
\eeq
Let $S$ be the union of all groups in $\cG^*$, i.e.
$S = \bigcup_{G \in \cG^*} G$, and the size of $S$ be $|S| = s$.
%
%
We are going to use three lemmas in order to prove the above bound. The first lemma, proved in~\cite{crpw12}, upper bounds the Gaussian width by an expected distance to the normal cone as follows.

\begin{lemm}[\cite{crpw12} Proposition 3.6]
Let $\C$ be any nonempty convex in $\R^p$, and $\g \sim \cN(0,I_p)$ be a random gaussian vector. Then
\beq
\omega(\C\cap \s^{p-1}) \leq \E_\g[\text{dist}(\g, \C^*)]~,
\eeq
where $\C^*$ is the polar cone of $\C$.
\end{lemm}

Note that $\cN_\cA$ is the polar cone of $\cT_\cA$ by definition. Therefore, using Jensen's inequality, we obtain
\beq
\omega(\cT_\cA \cap \s^{p-1})^2 \leq \E_\g^2[\text{dist}(\g, \cN_\cA)] \leq \E_\g[\text{dist}(\g, \cN_\cA)^2] \leq \E_\g[\|\g -  \z(\g)\|_2^2]~,
\eeq
where $\z(\g) \in \cN_\cA$ is a (random) vector constructed to lie always in the normal cone. The construction proceeds as follows.

\noindent{\bf Constructing $\mathbf{\z(\g)}$:}
Note that $\btheta^*_{S^c} = 0$. Let us choose a vector $\v \in \cN_\cA$ such that
\beq
\ksupd{\v} = 1~\text{and}~\v_{S^c} = 0~.
\eeq
We can choose an appropriately scaled $\v$ so that
\beq
\langle \v,\btheta^*\rangle = \ksup{\btheta^*}~,
\eeq
and let us write without loss of generality $\v = [\v_S ~~\v_{S^c}]$.

Next, let $\g \sim \cN(0,I_p)$, and write $\g = [\g_S~~ \g_{S^c}]$. We define the quantity
\beq
t(\g)  = \max \left\{ \|\g_G\|_2~:~G \in \cG^{(k)}, G \subseteq S^c\right\} = \max \left\{ \left( \sum_{i \in G}\g_i^2\right)^{\frac{1}{2}}~:~G \in \cG^{(k)}, G \subseteq S^c\right\}~,
\eeq
and let $\z = \z(\g) = [\z_S~~\z_{S^c}]$ such that
\beq
\z_S = t(\g) \v_S, ~~~~ \z_{S^c} = \g_{S^c}~.
\eeq
Note that
\beq
\langle \z, \btheta^* \rangle = t(\g) \langle\v_S, \btheta^*_S\rangle = t(\g) \ksup{\btheta^*}~,
\eeq
and
\begin{align}
\ksupd{\z} &= \max\left\{ \|\z_G\|_2~:~G\in \cG^{(k)}\right\}\\
            & = \max\Big\{  \max\{\|\z_G\|_2~:~G \in \cG^{(k)}, G \subseteq S \}~,~\max\{\|\z_G\|_2~:~G \in \cG^{(k)}, G \subseteq S^c \}   \Big\}\\
            & \stackrel{(a)}{=} \max\Big\{  t(\g)\ksupd{\v}~, ~ t(\g) \Big\}\\
            & = t(\g)
\end{align}
where $(a)$ follows from the definition of $t(\g)$ and the fact that
\beq
\max\{\|\z_G\|_2~:~G \in \cG^{(k)}, G \subseteq S \} = t(\g) \max\{\|\v_G\|_2~:~G \in \cG^{(k)}, G \subseteq S \} = t(\g)\ksupd{\v}~,
\eeq
and since $\ksupd{\v} =1$. Therefore, $\z(\g) \in \cN_\cA(\btheta^*)$ by definition in~\eqref{eq:normal cone} .

In order to upper bound the expectation of $t(\g)$, we use the following comparison inequality from~\cite{rarn12}.
\begin{lemm}[\cite{rarn12} Lemma 3.2]
\label{Mlem:maxbound}
Let $q_1,q_2,\ldots,q_L$ be $L$, $\chi$-squared random variables with $d$ degrees of freedom. Then
\beq
\E\left[\max_{1\leq i \leq L} q_i\right] \leq \left(\sqrt{2\log L} + \sqrt{d}\right)^2~.
\eeq
\end{lemm}

Last, we prove an upper bound on the expected value of $t(\g)$, as shown in the following lemma.
\begin{lemm}
Consider $\cG^* \subseteq \cG^{(k)}$ to be the set of groups intersecting with the support of $\btheta^*$, and let $S$ be the union of groups in $\cG^*$, such that $s = |S|$. Then,
\beq
\E_\g[t(\g)^2] \leq \left(\sqrt{2k\log \left(p - k - \ceil{\frac{s}{k}} +2\right)} + \sqrt{k}\right)^2~.
\eeq
\end{lemm}

\proof
Note that
\begin{align}
\E_\g[t(\g)^2] & = \E_\g\left[ \left(\max \left\{ \|\g_G\|_2~:~G \in \cG^{(k)}, G \subseteq S^c\right\}\right)^2\right] \\
& \label{eq:maxg bnd} \leq \E_\g\Big[  \max \left\{ \|\g_G\|_2^2~:~G \in \cG^{(k)}, G \subseteq S^c\right\}\Big]
\end{align}

Each term $\|\g_G\|_2^2$ is a $\chi$-squared variable with at most $k$ degrees of freedom. Since the set $S$ has size $s$, the set $\cG^*$ has to contain at least $s_k=\ceil{\frac{s}{k}}$ groups of size $k$. Therefore,
\beq
s = |S| \geq k + (s_k -1)~,
\eeq
and therefore the size of its complement is upper bounded by
\beq
|S^c| \leq p - k - s_k +1~.
\eeq

Therefore the following inequality provides an upper bound on the number of groups involved in computing the maximum in~\eqref{eq:maxg bnd}
\begin{align}
\Big|\Big\{G \in \cG^{(k)}, G \subseteq S^c\Big\} \Big| &\leq {p - k - s_k +1 \choose k} + {p - k - s_k +1 \choose k -1 } + \cdots + {p - k - s_k +1 \choose 1} \\
& \leq (p - k - s_k +2)^k
\end{align}
where we have used the following inequality
\beq
{n \choose h } \leq \frac{n^h}{h! },~~\forall n\geq h \geq 0~,
\eeq
which also provides
\beq
\sum_{h=1}^k {n \choose h} \leq (n+1)^k~.
\eeq
Therefore, we can upper bound~\eqref{eq:maxg bnd} using Lemma~\ref{Mlem:maxbound} as
\begin{align}
\E_\g[t(\g)^2] & \leq \E_\g\Big[  \max \left\{ \|\g_G\|_2^2~:~G \in \cG^{(k)}, G \subseteq S^c\right\}\Big] \\
&\leq \left(\sqrt{2\log \left((p - k - \ceil{\frac{s}{k}} +2)^k \right)} + \sqrt{k}\right)^2
\end{align}
and the statement follows. \qed

Now we are ready to prove the upper bound on the Gaussian width. First, note that
\begin{align}
\omega(\cT_\cA(\btheta^*) \cap \s^{p-1})^2 &\leq \E_\g[\text{dist}(\g, \cN_\cA(\btheta^*))^2] \\
& \substack{(a) \\ \leq} ~\E_\g[\|\g -  \z(\g)\|_2^2] \\
& = \E_\w[\|\z_S - \g_S\|_2^2]\\
&\substack{(b) \\ = }~~ \E[\|\z_S\|_2^2] + \E[\|\g_S\|_2^2] \\
& ~\substack{(c)\\=}~ \E[t(\g)^2]\cdot\|\v_S\|_2^2 + |S|\\
& ~\substack{(d)\\\leq}~ \left(\sqrt{2k\log \left((p - k - \ceil{\frac{s}{k}} +2) \right)}  + \sqrt{k}\right)^2 \cdot \ceil{\frac{s}{k}} + s~,
\end{align}
where $(a)$ follows from the definition of distance to a set, $(b)$ follows from the independence of $\g_S$ and $\g_{S^c}$, $(c)$ follows from the fact that the expected length of an $|S|$ length random i.i.d. Gaussian vector is $\sqrt{|S|}$, and $(d)$ follows since $|S| = \frac{k s}{k} $, and that $\|\v_S\|_2 \leq \sqrt{\ceil{\frac{s}{k}}} \ksupd{\v_S} = \sqrt{\ceil{\frac{s}{k}}}$. Thus inequality~\eqref{eq:ksup width} follows.
\qed

Next, we prove inequality~\eqref{eq:ksup lambda_p}.
Let us denote $\t  =\bX^T \left(\frac{\w}{\|\w\|_2}\right)$, and note that $\t \sim \cN(0,I_p)$. Also note that
$\E\left[\cR^*(\bX^T \w)\right] = E[\|\w\|_2\|] \E[\cR^*(\t)]$, and
\begin{align}
 \ksupd{\t} = \max\{\|\t_G\|_2~:~G\in\cG^{(k)}\}~.
\end{align}
Therefore, we can use Lemma~\ref{Mlem:maxbound} in order to bound the expectation $\E[\ksupd{\t}]$ as
\begin{align}
\E[\ksupd{\t}] &= \E[\max\{\|\t_G\|_2~:~G\in\cG^{(k)}\}] \\
&= \E[\max\{\|\t_G\|_2~:~G\in\cG^{(k)}, ~|G| = k\} \\
&\leq \left(\sqrt{2\log {p\choose k}} + \sqrt{k}\right) \\
&\leq \left(\sqrt{2k\log\left(\frac{pe}{k}\right) } + \sqrt{k}\right)~,
\end{align}
where we have used the following inequality obtained using Stirling's approximation
\beq
{p\choose k} \leq \left(\frac{pe}{k}\right)^k ~.
\eeq
Therefore, inequality~\eqref{eq:ksup lambda_p} follows, and by our choice of $\lambda_p$, with high probability, $\btheta^*$ lies in the feasible set.

Last, note that 
\beq
\omega(\Omega_\cR) =  \E[\ksupd{\t}] \leq \left(\sqrt{2k\log\left(\frac{pe}{k}\right)} + \sqrt{k}\right)~,
\eeq
as proved above.
\qed

\end{document}